\documentclass[11pt]{article}
\usepackage{acl}
\usepackage{subcaption}

\usepackage{times}
\usepackage{latexsym}
\usepackage[T1]{fontenc}
\usepackage[utf8]{inputenc}
\usepackage{microtype}
\usepackage{inconsolata}
\usepackage{graphicx}
\usepackage{amsmath}
\usepackage{booktabs}
\usepackage{enumitem}
\usepackage{makecell}
\usepackage{pifont} 
\usepackage{multirow}
\usepackage{algorithm}
\usepackage{algpseudocode}
\usepackage{xcolor}

\title{InsightChain: Optimized Chain-of-Insight Analytics for LLM-driven Data Visualization}

\author{Hanya Sun$^{\ast}$ \quad Chen Zhang$^{\ast\dagger}$ \quad \textbf{Sheng Liang} \quad \textbf{Yongyue Zhang} \quad \textbf{Yong Liu}     \\
  Huawei Technologies Co., Ltd. \\
  \tt{zhang.chen4@huawei.com}
}

\begin{document}
\maketitle

\begingroup
\renewcommand{\thefootnote}{\fnsymbol{footnote}}

\footnotetext[1]{Co-first Authors}
\footnotetext[2]{Corresponding Author}

\endgroup

\begin{abstract}
Large language models (LLMs) are increasingly used for automated data visualization, yet existing approaches often frame visualization generation as a single-step mapping from user query to figure or code, overlooking the iterative analytical reasoning process of expert analysts. We present InsightChain, a four-stage visualization prompting pipeline (Explore--Focus--Test--Present) that emulates expert analytical workflows, together with VG-COPRO, a vision-guided automatic prompt optimization (APO) method adapted to jointly optimize such multi-stage, executable pipelines. To address the evaluation gap for complex data visualization, we introduce the Insight Progression Metric (IPM), a rubric combining four text-based dimensions with a vision-based dimension. We assess IPM through a 100-chain human pilot and an expanded 300-chain agent-based evaluation spanning all ten domains. Experiments on public datasets show that InsightChain consistently outperforms competing prompting baselines. Existing APO methods fail to yield consistent gains on this multi-stage task, whereas VG-COPRO improves performance in both in-domain and cross-domain settings.


\end{abstract}


\section{Introduction}
\label{sec:intro}

Large language models (LLMs) have significantly advanced automated data visualization \citep{dibia-2023-lida, tian2024chartgpt}. However, most existing approaches treat visualization as a single-step mapping from query to executable code or rendered figure, bypassing intermediate validation \citep{wang2023chatgpt}. This one-shot approach contradicts expert sensemaking workflows \citep{pirolli2005sensemaking}, which are inherently iterative: analysts explore broad data distributions \citep{tukey1977exploratory}, formulate specific hypotheses \citep{heer2012interactive}, and refine visual encodings \citep{munzner2014visualization} to synthesize reliable insights \citep{sacha2014knowledge}. By compressing this multi-stage reasoning into a single run, current LLM strategies often yield superficial visualizations that lack statistical rigor and fail to uncover deep insights in complex real-world data.

To bridge this gap, we introduce InsightChain, a multi-stage prompting pipeline that emulates expert workflows through four stages: Explore, Focus, Test, and Present. Because these stages are interdependent, optimizing their prompts requires feedback that accounts for downstream execution and the rendered output. We therefore study how existing automatic prompt optimization (APO) methods transfer to this setting and introduce VG-COPRO, a task-adapted method that combines COPRO's per-predictor search with OPRO-style proposal and vision-aware, pipeline-level evaluation.

Evaluating this form of multi-stage analysis is difficult because a table may support multiple valid analytical directions, leaving no unique reference insight or visualization, while final-output metrics cannot determine whether intermediate hypotheses are grounded in executed evidence. We therefore introduce the Insight Progression Metric (IPM), a rubric tailored to multi-step visualization tasks, for evaluating both the analytical trajectory and the final visualization. Our experiments address four questions: (1) how well IPM agrees with human evaluation, and how robustly it behaves across repeated and cross-family judgments; (2) whether the chain pipeline improves over non-chain prompting baselines across task models; (3) whether VG-COPRO yields consistent gains over the unoptimized chain and existing APO methods; and (4) whether the optimized prompts transfer when the Present stage is rendered by a text-to-image model. Our contributions are three-fold:

\begin{itemize}
    \item We propose \textbf{InsightChain}, an execution-grounded prompting pipeline for automated data visualization that follows four stages of expert sensemaking: Explore, Focus, Test, and Present.

    \item We systematically study how existing APO methods transfer to multi-stage, executable data analysis and introduce \textbf{VG-COPRO}, a task-adapted optimizer that combines dimension-level diagnostic feedback, a leave-one-domain-out robustness objective, and vision-aware evaluation.

    \item We introduce the \textbf{Insight Progression Metric (IPM)}, a rubric for evaluating multi-step visualization tasks, assess its agreement with human ratings, and examine its repeatability and cross-family robustness through agent-based judgments. Experiments across ten domains demonstrate the effectiveness of InsightChain and VG-COPRO, including transfer to a text-to-image renderer.

\end{itemize}


\section{Related Work}
\label{sec:related-work}

\subsection{LLM-driven visualization}
\label{sec:rw-viz}
Early systems for natural-language-driven chart generation typically treat the task as a single-pass mapping from query to code or chart specification. For instance, while LIDA \citep{dibia-2023-lida} introduces an LLM-backed goal-then-code pipeline, its goal generation and code generation remain decoupled, exchanging no execution-grounded context. Similarly, ChartGPT \citep{tian2024chartgpt} relies on a single prompt conditioned on user queries and schemas, and Prompt4Vis \citep{li2025prompt4vis} improves few-shot selection but rigidly retains the one-shot input--output structure. Other systems, such as Chat2VIS \citep{maddigan2023chat2vis} and multi-agent extensions like Data-to-Dashboard \citep{wang2025data2dashboard}, add role decomposition, yet still lack an execution-grounded insight-progression objective. These systems primarily address data-plus-intent-to-visualization and remain the closest executable visualization baselines available. InsightChain instead begins without a predefined analytical intent and links iterative reasoning with execution feedback; our comparison therefore measures their practical performance under this open-ended setting rather than assuming identical task formulations.

\subsection{Insight discovery and analytical report generation}
\label{sec:rw-insight}

InsightBench \citep{sahu2024insightbench} and DiscoveryBench \citep{majumder2024discoverybench} evaluate generated textual insights or hypotheses against reference discoveries. DAComp \citep{dacomp2025} evaluates analytical reports conditioned on predefined business questions, while Multimodal DeepResearcher \citep{yang2025multimodaldeepresearcher} begins with a topic and web retrieval to produce an interleaved multimodal report. These tasks are related to InsightChain in analytical breadth but differ in their inputs and outputs: InsightChain begins with a single table and no predefined intent, and requires executable exploration culminating in a visualization. We therefore discuss these systems as related tasks but do not directly compare scores obtained under their incompatible evaluation protocols.

\subsection{Visualization quality metrics}
\label{sec:rw-metrics}
Existing evaluation metrics for LLM-driven data visualizations generally fall into three categories, each exhibiting structural limitations that motivate the design of our IPM rubric.

Specification-based metrics, such as nvBench \citep{luo2021nvbench} and Prompt4Vis \citep{li2025prompt4vis}, judge a chart as correct only if its Vega-Lite specification matches a reference exactly, penalizing equally valid visualizations with different encodings. Text-based metrics, like Chart-to-Text \citep{kantharaj2022c2t}, compare generated captions with references using BLEU; they tolerate surface differences but ignore whether the chart itself is accurate or renderable. Vision-language metrics, including CLIP similarity \citep{radford2021learningtransferablevisualmodels}, LIDA \citep{dibia-2023-lida}, and chart-VQA adaptations, evaluate rendered images directly, yet mainly capture image--text alignment rather than data fidelity, so charts with misleading axes or hidden aggregations may score higher than honest representations.

Our IPM addresses these limitations by evaluating both the intermediate analytical reasoning and the visual fidelity of the outputs. A detailed property comparison is provided in Table~\ref{tab:metrics-compare} (Appendix~\ref{sec:appendix-metrics-compare}).

\subsection{Automatic prompt optimization}
\label{sec:rw-apo}
Recent work on APO spans several algorithmic paradigms: Bayesian search \citep[MIPROv2,][]{opsahl2024miprov2}; coordinate ascent \citep[COPRO,][]{khattab2024dspy}; LLM-as-optimizer \citep[OPRO,][]{yang2024opro}; evolutionary search \citep[EvoPrompt,][]{guo2024evoprompt}; and fixed structured-template scaffolds \citep[Plan-and-Solve,][]{wang2023ps}. These paradigms are typically evaluated on text-only classification or QA benchmarks (HotpotQA, GSM8K, IRIS) with large validation sets and discrete correctness labels. However, applying them to multi-step visualization agents reveals two critical structural limitations: they optimize prompts in isolation rather than as a pipeline, and they rely solely on text-based signals, remaining blind to the visual fidelity of the rendered figure. These limitations motivate our design of VG-COPRO, a vision-guided, pipeline-aware approach.



\begin{figure*}[htbp]
    \centering
    \includegraphics[width=0.9\textwidth]{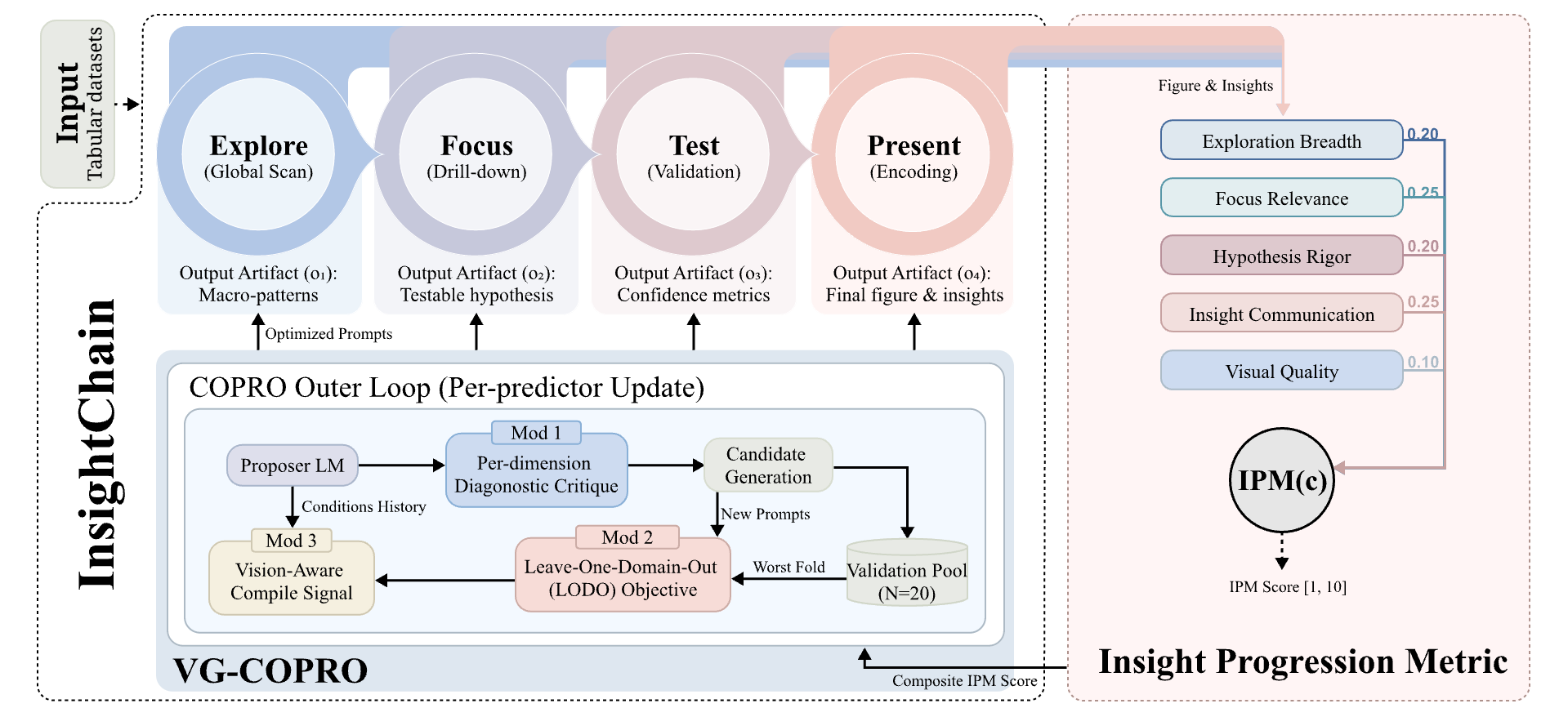}
    \caption{\textbf{System overview.}
        \textbf{InsightChain} (left) consists of (i) a four-stage Explore$\to$Focus$\to$Test$\to$Present pipeline, ingesting tabular datasets and emitting an output artefact ($o_t$) at each stage that conditions the next; and (ii) VG-COPRO, the prompt optimizer that wraps an OPRO-style proposer LM in COPRO's per-predictor outer loop, with three modifications selecting candidate prompts on a 20-example validation pool by composite IPM score.
        \textbf{IPM} (right): a five-dimensional rubric with weights $(0.20, 0.25, 0.20, 0.25, 0.10)$ scored by an LLM-as-judge, producing a composite score that serves both as the evaluation metric and as the VG-COPRO compile signal.
    }
    \label{fig:insightchain_overview}
\end{figure*}

\section{Methodology}
\label{sec:methodology}

\subsection{The InsightChain Pipeline}
\label{sec:framework}

\paragraph{Task formulation.}
Given a single tabular dataset $D$ without a predefined analytical question or visualization intent, the system must identify a worthwhile analytical direction, test the resulting hypothesis through executable analysis, and communicate the supported finding in a visualization. The output is therefore a multi-stage reasoning-and-execution trace together with a final figure, rather than a chart generated from a user-specified intent. InsightChain implements this task through a four-stage pipeline and uses VG-COPRO to optimize its stage-specific prompts.

Visualization analysis is formalized as a structured four-stage pipeline for iterative reasoning, $\mathcal{S} = (t_1, t_2, t_3, t_4)$\footnote{The four stages $t_1, t_2, t_3$, and $t_4$ correspond to the Explore, Focus, Test, and Present phases, respectively. Explore corresponds to Tukey's exploratory data analysis \citep{tukey1977exploratory}; Focus to the hypothesis-formulation step of interactive analysis \citep{heer2012interactive}; Test to standard statistical validation; and Present to the encoding-and-communication phase of visualization design \citep{munzner2014visualization}.}. At each stage $t$, a predictor $\pi_t$, implemented as a DSPy chain-of-thought module with a typed signature, produces a structured output $o_t = \pi_t(x_t; \theta_t)$, where $x_t$ denotes the input context and $\theta_t$ denotes the stage-specific prompt.

Each output $o_t$ contains an executable Python program $\mathrm{py}_t$,
which is executed by a sandboxed executor $\mathcal{E}$ with retry budget
$B = 2$:
\begin{equation}
\label{eq:exec}
e_t = \mathcal{E}^{(B)}(\mathrm{py}_t, D)
\end{equation}
where $D$ denotes the input table. Upon execution, the environment returns an execution state $e_t = (\mathrm{stdout}_t, \mathrm{figs}_t, \mathrm{err}_t)$. Here, $\mathrm{stdout}_t$ captures the standard textual output, $\mathrm{figs}_t$ denotes the rendered visualizations, and $\mathrm{err}_t$ logs runtime errors. If $\mathrm{err}_t \neq \emptyset$, the error trace is fed back to the predictor $\pi_t$, guiding the regeneration of $o_t$ and thus forming a self-correcting execution loop.

Downstream stages condition on prior outputs and execution traces. We collect
the record of each completed stage into a running context
$C_{<t} = (o_1, e_1, \ldots, o_{t-1}, e_{t-1})$, with $C_{<1} = \emptyset$.
The input to every predictor is uniformly
\begin{equation}
\label{eq:inputs}
x_t = (\sigma(D),\; C_{<t}),
\end{equation}
where $\sigma(D)$ is a fixed dataset profile (column names, dtypes, summary
statistics, and a five-row sample). The pipeline's final deliverable is
$(o_4.\textsc{final\_insight},\, \mathrm{figs}_4)$; the stage records
$\{(o_t, e_t)\}$ serve as the evaluation basis for IPM and the optimization signal for VG-COPRO. The prompt $\theta_t$ operates on top of the typed signature. We present the Explore instruction verbatim below to illustrate the prompt format (full-stage prompts are provided in Appendix~\ref{sec:appendix-algorithm}):
\begin{quote}\small\itshape
``Given a dataset summary and sample, generate Python visualization code that
broadly explores the data landscape with 2--3 overview charts. Identify
distributions, relationships, temporal patterns, and outliers. Print key
observations and identify non-obvious patterns worth investigating.''
\end{quote}

\subsection{Insight Progression Metric (IPM)}
\label{sec:ipm}
Before diving into the details of prompt optimization, we first define the optimization objective used to quantify improvements in insight visualization. We propose the Insight Progression Metric (IPM), a weighted linear combination of five quality dimensions computed from the chain’s source code, intermediate insights, and rendered figures:
\begin{equation}
\mathrm{IPM}(c) = \sum_{d \in \mathcal{D}} w_d \cdot \phi_d(c),
\end{equation}
where $\mathcal{D} = \{$\textsc{Exploration Breadth}, \textsc{Focus Relevance}, \textsc{Hypothesis Rigor}, \textsc{Insight Communication}, \textsc{Visual Quality}$\}$, $\phi_d \in [1, 10]$ is the LLM-as-judge score, and $w = (0.20, 0.25, 0.20, 0.25, 0.10)$\footnote{The weights were specified manually from the task requirements. Section~\ref{sec:ipm-validation} evaluates uniform, perturbed, fully random, and data-fitted alternatives and shows that the principal conclusions are stable across these choices.}. 

The first four dimensions correspond to the four stages of InsightChain and are evaluated by a text-based judge using the generated code and insight descriptions from each stage. \textsc{Visual Quality} is assessed by a vision-language model using the figure rendered in the Present stage.

\textsc{Visual Quality} contributes $10\%$ to the composite IPM score; it complements the four text-based dimensions rather than determining the overall evaluation on its own. Within this dimension, the central criterion is insight embodiment: whether the visualization effectively communicates the final finding through an appropriate chart type and elements such as annotations, reference lines, and statistical overlays, rather than merely displaying the data correctly. A faithful but uninformative chart therefore receives only a moderate or low score. The score is capped at $3$ if either (1) the figure misrepresents the data or (2) fundamental legibility failures trigger the display gate. Detailed gate conditions appear in the annotator handbook (Appendix~\ref{sec:appendix-annotator}). If the final figure fails to render, the score is set to $\phi_5=1$. All results in §~\ref{sec:results} use this IPM formulation, which also provides the optimization signal for the APO methods.
 
\subsection{Prompt Optimization Paradigms}
\label{sec:apo}

We study automatic prompt optimization (APO) for InsightChain, treating proprietary LLMs as fixed-weight black-box APIs. First, we examine how existing APO methods interact with our four-stage InsightChain framework. We recompile the same InsightChain using five representative methods spanning the current APO taxonomy: MIPROv2 \citep{opsahl2024miprov2}, COPRO \citep{khattab2024dspy}, OPRO \citep{yang2024opro}, EvoPrompt \citep{guo2024evoprompt}, and Plan-and-Solve \citep{wang2023ps}. These algorithms differ on two axes, search target and proposer mechanism, as summarised in Table~\ref{tab:apo} of Appendix~\ref{sec:appendix-apo-design}.

Because OPRO, EvoPrompt, MIPROv2, and Plan-and-Solve were originally designed for single-prompt optimization, while InsightChain consists of four predictors, we adapt all five methods using COPRO’s per-predictor outer-loop strategy. Specifically, the instruction of one stage is optimized at a time while the remaining stages are kept fixed. Only the inner proposal-generation mechanism varies for different APOs. Moreover, Plan-and-Solve serves as a no-search baseline. Its proposer slot is replaced with a fixed plan-then-solve template applied once per predictor, allowing us to isolate the contribution of iterative search beyond a carefully designed template.

\subsection{VG-COPRO: a vision-guided APO}
\label{sec:vg-copro}



When adapted to InsightChain using the common per-predictor outer loop, existing APO methods yield no consistent improvement across both evaluation splits. This setting presents three task-specific difficulties: a prompt change at one stage propagates through downstream execution, average validation performance may not transfer across domains, and text-only feedback does not assess the rendered figure. We therefore develop VG-COPRO as a task-adapted optimizer built on COPRO and OPRO, rather than as a new general-purpose APO framework. Each candidate stage prompt is assessed by executing the complete pipeline, allowing optimization to account for downstream effects, cross-domain robustness, and visual feedback.

VG-COPRO is built upon the COPRO outer loop for iterative per-predictor optimization. For the inner loop, we adopt OPRO's proposer, since it is the only method achieving a positive in-domain gain and supporting non-scalar feedback in Table~\ref{tab:baselines}. Based on this structure, we introduce three independent modifications to VG-COPRO.

\textbf{Mod 1: per-dimension diagnostic critique.} Instead of a single scalar reward, the proposer LM receives the full five-dimensional IPM breakdown together with the weakest-performing dimension. This converts the optimization feedback into a structured diagnostic signal, extending the textual-feedback paradigm of \citet{yuksekgonul2024textgrad}. \textbf{Mod 2: leave-one-domain-out (LODO) objective.} Fitness is defined as the minimum IPM across the five disjoint in-domain folds, rather than their mean. This formulation acts as a worst-case robustness objective, inspired by Group DRO \citep{sagawa2020groupdro}, and better aligns the optimization signal with cross-domain generalization. \textbf{Mod 3: vision-aware compile signal.} Every candidate prompt is evaluated using the full 5-dimensional IPM, including the vision-judge assessment. The full pseudocode is shown in Algorithm~\ref{alg:vg-copro} (Appendix~\ref{sec:appendix-algorithm}).

\section{Experimental Setup}
\label{sec:setup}


\paragraph{Datasets.}
We aggregate $285$ tabular datasets spanning $10$ domains from seven public sources. Under a leave-five-domains-out protocol, five domains supply $150$ datasets for compilation (train $n{=}95$, validation $n{=}20$, and in-domain test $n{=}35$), while the other five domains provide the cross-domain test set ($n{=}135$). Each $n$ denotes an independent analysis task associated with a distinct dataset, rather than a table row. The median dataset contains approximately $6{,}000$ rows; the in-domain and cross-domain test sets contain approximately $550$K and $2.09$M records, respectively. Overall, $91.9\%$ of the datasets come from real public sources, with a small synthetic-template fallback for underrepresented demographics and energy domains. Source composition, domain counts, and split sizes are detailed in Appendix~\ref{sec:appendix-datasets}.

\paragraph{IPM judges.}
The primary text and visual judge used for IPM scoring is Gemini~3~Flash, with inference temperature set to $0$. To assess potential same-family judge bias, GPT-4.1 and Qwen-VL-Max re-score a stratified subsample of $n=25$ chains per condition across three baselines and five APO methods using the same rubric.

\paragraph{Task models and other metrics.}
Chain generation and baselines run on Gemini~3~Flash (primary; temperature $0.3$, $10{,}000$-token output limit) and on GPT-4.1 for cross-model replication. Generated code executes in a sandboxed subprocess (60~s timeout, up to two retries). For the auxiliary visualization metrics, we use OpenCLIP ViT-L/14 for CLIP similarity, SacreBLEU for Chart-to-Text BLEU, and the installed lida v0.0.14 package for the LIDA VLM evaluator.

\paragraph{APO compilation.}

All APOs share a single train/val pool ($n_{\text{val}}=20$) drawn from the five training domains, with a per-predictor budget of three compile iterations. VG-COPRO evaluates each candidate using five leave-one-domain-out folds constructed exclusively from the training domains: finance, healthcare, social science, climate, and sports. The optimizer never accesses the five cross-domain test domains: education, transportation, e-commerce, energy, and demographics. 


\paragraph{Text-to-Image transferability.}
We also evaluate the transferability of InsightChain prompts optimized with VG-COPRO to Text-to-Image (T2I) models. In this setting, the Present stage is rendered using \texttt{gemini-3.1-flash-image-preview}, while the three preceding stages, datasets, IPM weights, and evaluation judges remain unchanged.

\section{Results}
\label{sec:results}

\subsection{IPM Validation}
\label{sec:ipm-validation}


We assess IPM using complementary tests of controlled-band discriminability, human agreement, weight sensitivity, judge self-consistency, dimension overlap, and cross-family robustness.
Extended analysis on agent-based repeatability, expert-audited criteria, and close-pair discrimination are stated in Appendix~\ref{sec:appendix-validation-protocols}.

For the controlled-band analysis, we construct a corpus of 100 chains from 20 held-out datasets, where each dataset is evaluated under five prompt regimes corresponding to different quality levels: \textsc{high}, \textsc{medium-a}, \textsc{medium-b}, \textsc{low}, and \textsc{random-permutation}. The \textsc{high} regime encourages expert-level statistical analysis and publication-ready visualizations, while \textsc{medium-a} and \textsc{medium-b} use the same analyst-level prompt under different random seeds to measure robustness against stochastic variation. The \textsc{low} regime contains minimal plotting instructions without insight requirements, and \textsc{random-permutation} serves as a content-broken control with unlabeled and semantically meaningless figures. This paired $20 \times 5$ design ensures that all regime comparisons are controlled at the dataset level. The full system prompts and the chain-generation script are documented in Appendix~\ref{sec:appendix-regimes}.

\begin{figure}[h]
\centering
\includegraphics[width=\linewidth]{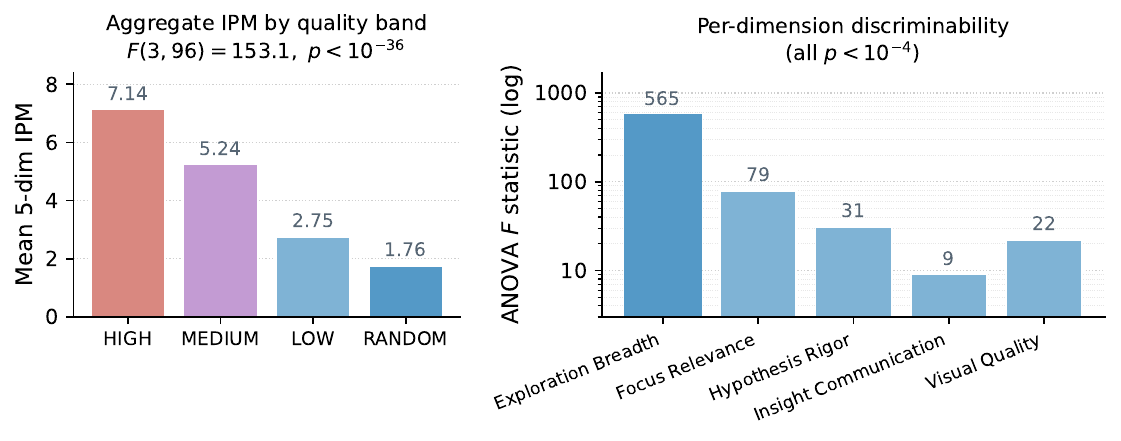}
\caption{IPM discriminability on the 100-chain pilot. \textbf{Left}: aggregate 5-dim IPM by quality band. \textbf{Right}: per-dimension ANOVA $F$ statistic (all $p < 10^{-4}$).}
\label{fig:discriminability}
\end{figure}

\paragraph{Discriminability.}
The five-dimensional IPM clearly distinguishes the controlled quality bands (Figure~\ref{fig:discriminability}). A one-way ANOVA conducted on the aggregate IPM scores across the four real-quality groups (\textsc{high}, \textsc{medium-a}, \textsc{medium-b}, and \textsc{low}) yields $F(3,96)=153.1$ with $p<10^{-36}$, indicating strong statistical separation between quality levels. As intended, IPM does not significantly distinguish \textsc{medium-a} from \textsc{medium-b}, which use the same prompt and differ only in random seed; it separates the remaining adjacent quality levels with Cohen's $d>1.8$. Each individual IPM dimension also independently separates the quality bands with statistical significance ($p<10^{-4}$).


\paragraph{Human agreement.}

Two members of the author team, trained using the annotator handbook and blind to both the chain source condition and the IPM scores, evaluated the 100-chain pilot (Appendix~\ref{sec:appendix-annotator}). Among the $n=97$ pilot chains with complete overall-quality ratings, the IPM composite score shows strong agreement with the human ratings, achieving Spearman $\rho=0.88$ (Table~\ref{tab:human-agreement} in Appendix~\ref{sec:appendix-ipm-validation}). IPM assigns systematically lower scores than the human annotators on the four text-based dimensions while closely matching them on \textsc{Visual Quality}. This difference primarily reflects a shift in scoring scale rather than disagreement in ordering, as IPM and the human annotators rank the quality regimes consistently.

\paragraph{Weight sensitivity.}
Because the IPM weights are manually specified, we test whether the conclusions depend on this choice. Uniform weights leave the ranking of all nine methods unchanged (Kendall's $\tau=1.0$), with all significance conclusions preserved. Across 500 small perturbations around the original weights, the rankings remain stable, with mean $\tau=0.98$ in-domain and $0.95$ cross-domain; VG-COPRO remains the top-performing method in almost every case. Under 500 completely random weight configurations, the advantage of chain-based methods over single-shot methods remains consistently significant. Data-fitted weights achieve nearly identical agreement to the manual weights ($\rho=0.894$ versus $0.882$). These results indicate that the central conclusions are not determined by the exact manual weighting.


\paragraph{Self-consistency.}

Because the main experiments use temperature-0 decoding with prompt caching, repeated evaluations are effectively deterministic. To estimate residual variability in the judges, we therefore conduct a self-consistency analysis by rescoring each pilot chain three independent times at temperature 0.5. For each evaluation dimension, self-consistency is computed as the mean pairwise agreement among the three resulting scores.

All five dimensions achieve Krippendorff’s $\alpha \geq 0.85$, exceeding the standard threshold for high reliability (Table~\ref{tab:judge-reliability}a, Appendix~\ref{sec:appendix-ipm-validation}).

\paragraph{Dimension decoupling.} 


On the 508 complete chains from the original three-condition evaluation (170 datasets evaluated with single-shot, uncompiled-chain, and LIDA baselines; two failed LIDA evaluations excluded), all text--vision correlations are below $0.15$, while the Focus--Hypothesis correlation is $0.50$ (Figure~\ref{fig:ipm-decoupling}). In the expanded analysis, Focus--Hypothesis correlations range from $0.57$ to $0.87$ across human and agent annotations, compared with $0.76$ for the primary judge. This pattern across datasets and annotators suggests that the correlation partly reflects the pipeline dependency whereby the Test stage builds on the direction established during Focus, while also indicating residual rubric overlap. When we merge \textsc{Focus} and \textsc{Hypothesis} into one dimension with a combined weight of $0.45$, the rankings of all nine methods remain unchanged in both settings (Kendall's $\tau=1.0$), with VG-COPRO ranked first. We retain the dimensions separately because their responses across controlled quality regimes provide distinct diagnostic information.

Moreover, Cronbach’s $\alpha$ for the five IPM dimensions remains well below the standard 0.7 threshold, indicating that the dimensions do not collapse into a single underlying factor. In particular, removing \textsc{Visual Quality} is the only modification that increases $\alpha$, suggesting that the visual dimension captures information not covered by the text-based dimensions. This supports our hypothesis that visualization quality is only weakly correlated with textual reasoning quality, motivating vision-guided evaluation and optimization.


\paragraph{Cross-family robustness.}

Because the Gemini judge belongs to the same model family as the task model, there is a potential risk of self-enhancement bias, where the judge may overrate generated outputs. To evaluate this possibility, we re-scored 25 chains per condition from three step-matched baselines and five published APO methods using two cross-family evaluators, GPT-4.1 and Qwen-VL-Max. The results show that Gemini assigns lower mean scores than both cross-family models (Table~\ref{tab:judge-reliability}b, Appendix~\ref{sec:appendix-ipm-validation}), while per-condition Kendall’s $\tau$ values remain non-negative across evaluators, suggesting the method comparisons reported in §~\ref{sec:apo-results} are unlikely to be artifacts of judge-model bias.


\begin{table*}[t]
\centering
\small
\setlength{\tabcolsep}{4pt}
\begin{tabular}{llcrrrcrrr}
\toprule
 & & \multicolumn{4}{c}{\textbf{In-domain (n=35)}} & \multicolumn{4}{c}{\textbf{Cross-domain (n=135)}} \\
\cmidrule(lr){3-6} \cmidrule(lr){7-10}
\textbf{Task model} & \textbf{Method} & IPM & $\Delta$ & $p_1$ & $d$ & IPM & $\Delta$ & $p_1$ & $d$ \\
\midrule
\multirow{9}{*}{Gemini~3~Flash}
 & \textbf{InsightChain (pre-APO)} & \textbf{7.24} & --      & --                     & --     & \textbf{7.38} & --      & --                     & --     \\
 \cmidrule(l){2-10}
 & Single-shot 4-prompt          & 6.05 & $-1.19$ & $1.7{\times}10^{-7}$   & $1.07$ & 6.38 & $-1.00$ & $1.6{\times}10^{-22}$  & $1.01$ \\
 & LIDA                          & 5.79 & $-1.46$ & $5.6{\times}10^{-10}$  & $1.40$ & 5.93 & $-1.45$ & $1.0{\times}10^{-23}$  & $1.05$ \\
 \cmidrule(l){2-10}
 & MIPROv2                       & 7.20 & $-0.05$ & $0.63$                 & --     & 7.15 & $-0.24$ & $1.00$                 & --     \\
 & COPRO                         & 7.08 & $-0.16$ & $0.82$                 & --     & 7.05 & $-0.33$ & $1.00$                 & --     \\
 & OPRO                          & 7.55 & $+0.30$ & $0.085$ & $0.24$ & 7.27 & $-0.11$ & $0.86$ & --     \\
 & EvoPrompt                     & 6.88 & $-0.37$ & $0.96$                 & --     & 7.00 & $-0.39$ & $1.00$                 & --     \\
 & Plan-and-Solve                & 7.07 & $-0.17$ & $0.90$                 & --     & 7.08 & $-0.31$ & $1.00$                 & --     \\
 & \textbf{VG-COPRO}      & \textbf{8.56} & \textbf{$+1.31$} & \textbf{$<\!10^{-5}$} & \textbf{$1.41$} & \textbf{8.54} & \textbf{$+1.16$} & \textbf{$<\!10^{-5}$} & \textbf{$1.06$} \\
\midrule
\multirow{3}{*}{GPT-4.1}
 & \textbf{InsightChain (pre-APO)} & \textbf{7.90} & --      & --                     & --     & \textbf{8.01} & --      & --                     & --     \\
 \cmidrule(l){2-10}
 & Single-shot 4-prompt          & 6.72 & $-1.18$ & $7.0{\times}10^{-6}$   & $0.86$ & 6.92 & $-1.09$ & $3.7{\times}10^{-18}$  & $0.86$ \\
 & LIDA                          & 4.29 & $-3.62$ & $9.4{\times}10^{-16}$  & $2.40$ & 4.33 & $-3.69$ & $2.5{\times}10^{-63}$  & $2.80$ \\
\bottomrule
\end{tabular}
\caption{Main-result table. $\Delta = \text{method} - \text{pre-APO}$ at the same task model; $p_1$ is a one-sided paired $t$-test; $d$ is the paired Cohen's effect size. Negative $\Delta$ means the pre-APO wins. Within each task model, rows are grouped as: (a) InsightChain before APO optimization, (b) non-chain baselines, and (c) InsightChain with APO.}
\label{tab:baselines}
\end{table*}

\subsection{Chain vs.\ Baselines}
\label{sec:baselines}
We compare InsightChain to step-matched non-chain baselines (Single-shot 4-prompt and LIDA) under two task models on an in-domain test ($n = 35$) and a cross-domain test ($n = 135$). 


\paragraph{InsightChain vs.\ step-matched baselines.}
InsightChain consistently outperforms the single-shot baseline across different task models (Table~\ref{tab:baselines}). The improvement margins on Gemini~3~Flash and GPT-4.1 are highly similar, suggesting that the gains mainly come from the multi-stage pipeline design rather than model-specific behavior. Both models also show a small positive cross-domain transfer gain, indicating that the pipeline generalizes well across domains.

\begin{figure*}[t]
\centering
\begin{minipage}[t]{0.32\textwidth}
\centering
\includegraphics[width=\linewidth]{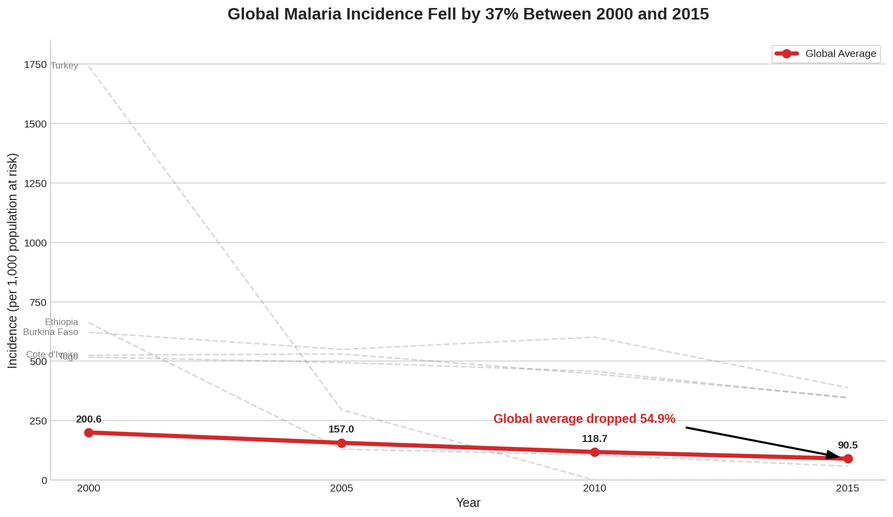}\\[2pt]
{\small (a) Single-shot\\IPM~$=6.60$,\ VQ~$=3$}
\end{minipage}\hfill
\begin{minipage}[t]{0.32\textwidth}
\centering
\includegraphics[width=\linewidth]{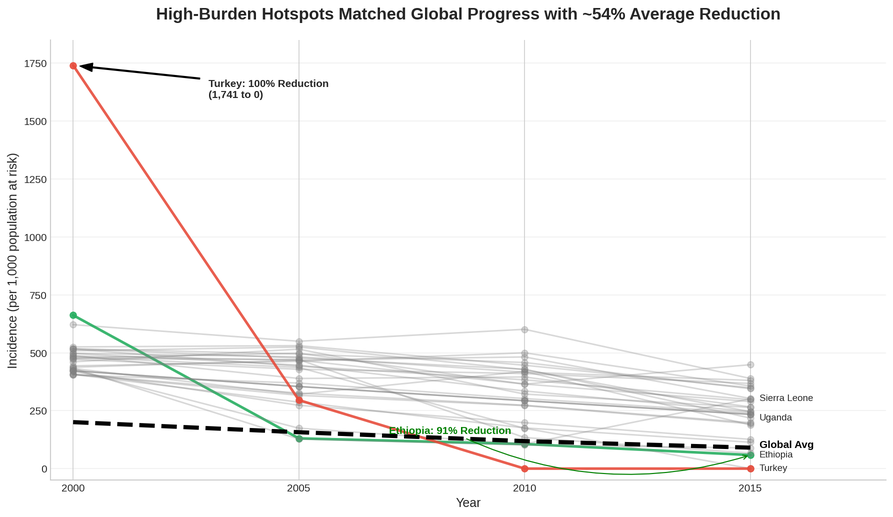}\\[2pt]
{\small (b) pre-APO \textsf{InsightChain}\\IPM~$=7.30$,\ VQ~$=9$}
\end{minipage}\hfill
\begin{minipage}[t]{0.32\textwidth}
\centering
\includegraphics[width=\linewidth]{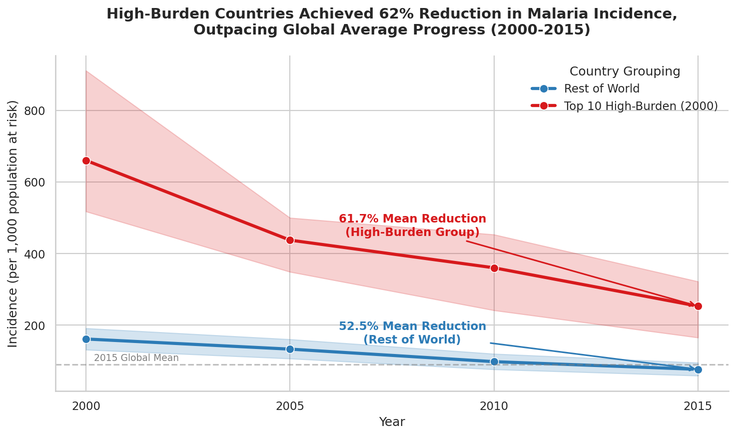}\\[2pt]
{\small (c) \textsf{InsightChain~$+$~VG-COPRO}\\IPM~$=9.10$,\ VQ~$=10$}
\end{minipage}
\caption{Qualitative comparison of visualizations on the \textsc{malaria\_inc} dataset. (a) Non-chain baseline: Produces a misleading chart where the title ($37\%$) contradicts the annotated value ($54.9\%$), triggering faithfulness-gate violation (VQ capped at 3). (b) InsightChain pipeline (pre-APO): Correct analytical pattern but poor layout and visual clutter. (c) InsightChain with VG-COPRO: A dual-trend plot with clear annotations and insights.}
\label{fig:malaria_example}
\end{figure*}


\paragraph{IPM vs Other Visualization Metrics}

We also re-scored every chain on the cross-domain test set using three published evaluators: CLIP image--caption similarity \citep{radford2021learningtransferablevisualmodels}, Chart-to-Text BLEU \citep{kantharaj2022c2t}, and the official LIDA VLM evaluator \citep{dibia-2023-lida}. Surprisingly, none of them captures the chain's insight advantage, yet this is understandable since they focus on surface-level alignment (CLIP), caption overlap (BLEU), or code correctness (LIDA), leading to saturation or floor effects and failing to distinguish deeper insights (Table~\ref{tab:robustness}a). This directly confirms the motivation for proposing IPM.

\begin{table}[h]
\centering
\small
\setlength{\tabcolsep}{2pt}
\begin{tabular}{lrrrrr}
\toprule
 & \textbf{Chain} & \textbf{Single} & \textbf{LIDA} & \textbf{$\Delta$} & \textbf{$d$} \\
\midrule
\multicolumn{6}{l}{\textbf{(a) Other Evaluation Metrics ($n=135$)}} \\
CLIP img--cap.\        & $0.305$ & $0.300$ & $0.300$ & $+0.00$ & $0.05$ \\
C2T BLEU               & $\sim\!0$ & $\sim\!0$ & $\sim\!0$ & --   & -- \\
LIDA VLM               & $9.38$  & $9.47$  & $9.38$  & $-0.10$ & $-0.19$ \\
\textbf{IPM (ours)}    & \textbf{7.36} & $6.32$ & $5.97$ & $\mathbf{+1.04}$ & $\mathbf{1.07}$ \\
\midrule
\multicolumn{6}{l}{\textbf{(b) Judge-free Code Validation ($n=133$)}} \\
$p$-value reports      & $\mathbf{1.13}$ & $0.26$ & $0.20$ & $+0.87$ & $0.43$ \\
\texttt{df} vars       & $\mathbf{6.19}$ & $5.19$ & $3.47$ & $+1.00$ & $0.42$ \\
\bottomrule
\end{tabular}
\caption{Results on cross-domain test set (means reported for InsightChain, Single-shot, and LIDA). $\Delta$ and $d$ measure the difference and effect size between InsightChain and Single-shot. \textbf{(a)} Three common visualization metrics fail to separate the methods, while IPM distinctly separates them ($d > 1$). \textbf{(b)} Two code metrics favor InsightChain ($p < 10^{-5}$).}
\label{tab:robustness}
\end{table}


\paragraph{Judge-free code validation.}
We further validate the IPM improvements using two judge-independent code metrics (Table~\ref{tab:robustness}b): the frequency of $p$-value reporting and the number of distinct dataframe variables analyzed. The former is extracted using regex matching over generated code, execution outputs, and narrative text, while the latter measures how broadly the pipeline explores the dataset schema through unique dataframe accesses. Both metrics consistently favor InsightChain over the single-shot baseline ($d \approx 0.4$, $p < 10^{-5}$), with even larger improvements over LIDA. These results suggest that InsightChain performs deeper statistical analysis and broader data exploration. In particular, the dedicated Test stage encourages explicit statistical testing and $p$-value reporting, while the Explore--Focus workflow promotes wider schema inspection before narrowing to specific hypotheses.


\subsection{APO on InsightChain}
\label{sec:apo-results}

All five published APO methods fail to improve performance on the cross-domain set, with results falling below the pre-APO baseline and showing no significant gains (Table~\ref{tab:baselines}). Among them, only OPRO achieves a positive in-domain improvement. We observe two key patterns. First, under limited optimization budgets, search-based APO methods provide little benefit: the no-search Plan-and-Solve baseline performs comparably to MIPROv2 and COPRO, suggesting that iterative search cannot outperform a strong fixed plan-and-execute scaffold under limited budgets. Second, score-blind proposers such as COPRO and EvoPrompt perform the worst. Although they generate prompts that appear more sophisticated, they often disrupt the underlying reasoning chain instead of improving it.

Appendix~\ref{sec:appendix-apo-dynamics} presents checkpoint-reconstructed optimization trajectories.  Within the three-iteration budget, these trajectories show that optimization can improve validation performance, but that the gains are stage- and method-dependent within our three-iteration budget.

These results reveal a key limitation of existing APO methods: they optimize solely based on textual artifacts from the reasoning chain. As a result, the optimized prompts often improve text-based metrics while failing to generate visually faithful or well-annotated figures. Overall, this suggests that the optimization signal should match the multimodal nature of the final evaluation, motivating the design of VG-COPRO.






\subsection{VG-COPRO on InsightChain}
\label{sec:vg-copro-results}

VG-COPRO improves IPM over the pre-APO baseline on both splits, with $p_1 < 10^{-5}$ and $d > 1$, and achieves the highest effect size among all APO methods examined, including the five APO methods (Table~\ref{tab:baselines}) and the three single-module ablations (Table~\ref{tab:apo-ablation}). It also surpasses the OPRO baseline by more than one IPM point on both splits. Figure~\ref{fig:malaria_example} illustrates this progression qualitatively on the cross-domain dataset \textsc{malaria\_inc}; Appendix~\ref{sec:appendix-full-chains} provides three additional stage-by-stage traces with dataset identifiers, statistical decisions, final figures, and per-dimension IPM scores.

To assess run-to-run stability, we evaluate three independent VG-COPRO compilations initialized with different seeds. Cross-domain improvements over the pre-APO chain are $+1.16$, $+0.44$, and $+0.91$ IPM points (mean $\pm$ standard deviation: $+0.84\pm0.37$), and all are statistically significant ($p\leq0.02$). In-domain improvements are likewise positive at $+1.31$, $+0.77$, and $+0.49$ (mean $\pm$ standard deviation: $+0.86\pm0.42$). Thus, although the magnitude varies across runs, the direction of improvement remains stable.

VG-COPRO's execution log provides stage-level diagnostic evidence. Under Mod~1, \textsc{Visual Quality} is the weakest dimension for the Focus, Test, and Present modules, while \textsc{Hypothesis Rigor} is weakest for Explore. Under Mod~2, the \textsc{sports} fold is most often the lowest-scoring LODO fold for Explore and Test and ties with \textsc{climate} for Focus; \textsc{climate} is most often weakest for Present.

These diagnostics steer the proposer toward prompts that prioritize insight-bearing annotations and appropriate visual encodings, rather than implementation-level Matplotlib detail that can improve text-judged artifacts without improving the rendered figure.

\paragraph{Ablation.}
We conduct an ablation experiment to attribute the contribution of each modification. As shown in Table~\ref{tab:apo-ablation}, each modification provides a significant improvement on both in-domain and cross-domain splits, so the three modifications are not redundant. Mod~2 (LODO) is the strongest single modification on mean $\Delta$, while the full combination yields the highest paired Cohen's $d$ on both splits, suggesting the effectiveness of our proposed three modifications in VG-COPRO.




\begin{table}[h]
\centering
\small
\setlength{\tabcolsep}{3pt}
\begin{tabular}{lrrrrrr}
\toprule
 & \multicolumn{3}{c}{\textbf{In-dom ($n{=}35$)}} & \multicolumn{3}{c}{\textbf{Cross-dom ($n{=}135$)}} \\
\cmidrule(lr){2-4} \cmidrule(lr){5-7}
\textbf{Variant} & IPM & $\Delta$ & $d$ & IPM & $\Delta$ & $d$ \\
\midrule
Hybrid         & $7.55$ & $+0.49^{\dagger}$ & $0.28$ & $7.27$ & $+0.24^{\dagger}$ & $0.19$ \\
+\,Mod 1       & $8.54$ & $+1.30$ & $1.06$ & $8.65$ & $+1.26$ & $0.84$ \\
+\,Mod 2       & $8.82$ & $\mathbf{+1.57}$ & $1.47$ & $8.74$ & $\mathbf{+1.36}$ & $1.06$ \\
+\,Mod 3       & $8.50$ & $+1.26$ & $0.93$ & $8.61$ & $+1.23$ & $0.95$ \\
\midrule
\textbf{VG-COPRO} & $\mathbf{8.74}$ & $+1.49$ & $\mathbf{1.62}$ & $\mathbf{8.63}$ & $+1.24$ & $\mathbf{1.15}$ \\
\bottomrule
\end{tabular}
\caption{Ablation of the three modifications for VG-COPRO (§~\ref{sec:vg-copro}). Hybrid baseline is based on the results of OPRO on chain pipeline. Each modification produces a significant improvement on both splits (all $p_1 < 10^{-5}$ for rows other than the Hybrid baseline). Mod~2 yields the largest mean $\Delta$; the full combination yields the largest paired Cohen's $d$, indicating lowest dispersion.}
\label{tab:apo-ablation}
\end{table}

\subsection{Generalization to Text-to-Image Models}
\label{sec:t2i}

A separate generalization question is whether the chain advantage persists when the Present stage is rendered by a text-to-image (T2I) model rather than a code model. To test this, we replace the Present-stage task model with a T2I variant. The chain outputs a 200–400-word visual brief which \texttt{gemini-3.1-flash-image-preview} renders directly, while the three reasoning stages retain their VG-COPRO-compiled instructions. The step-matched single-shot T2I baseline maintains the same four-prompt single-shot architecture as in §~\ref{sec:baselines}, with each prompt using the T2I Present-stage task model. Results are summarised in Table~\ref{tab:t2i}. The performance improvement ($\Delta$) reported in Table~\ref{tab:t2i} is similar to that observed in Table~\ref{tab:baselines}. This suggests that the benefits of InsightChain mainly come from its multi-stage chain-and-feedback workflow itself, rather than from a specific rendering backend such as code-based plotting or T2I generation.

\begin{table}[h]
\centering
\small
\setlength{\tabcolsep}{4pt}
\begin{tabular}{lrrrrr}
\toprule
\textbf{Split} & \textbf{n} & \textbf{Chain} & \textbf{Single-shot} & \textbf{$\Delta$} & \textbf{$d$} \\
\midrule
In-domain    & 35  & $8.04$ & $5.45$ & $+2.59$ & $2.17$ \\
Cross-domain & 135 & $7.60$ & $5.81$ & $+1.79$ & $1.50$ \\
\bottomrule
\end{tabular}
\caption{T2I transferability: mean IPM of InsightChain (T2I~Present) vs.\ single-shot~T2I baseline on the full test set. $n$ refers to the number of sampled datasets. All $p_1 < 10^{-14}$ by one-sided paired $t$-test.}
\label{tab:t2i}
\end{table}






\section{Conclusion}
\label{sec:conclusion}

We introduce InsightChain, an execution-grounded four-stage pipeline for open-ended insight discovery and visualization, together with IPM and the vision-guided optimizer VG-COPRO. The main experiments show that IPM separates controlled quality levels and agrees with human judgments, while InsightChain outperforms the included step-matched baselines across task models, with judge-free indicators supporting deeper analysis. Existing APO methods provide no consistent gain across both evaluation splits; VG-COPRO improves the chain in both settings, and the optimized pipeline retains its advantage with text-to-image rendering.



\section*{Limitations}
\label{sec:limitations}

\paragraph{IPM construct validity.}
Full five-dimensional human ratings are limited to the 100-chain pilot. The expanded 300-chain evaluation relies on repeated agent-based ratings and one expert's audit of three binary criteria. This evidence supports metric repeatability and separation of broad quality regimes, but does not fully validate every individual dimension or IPM's ability to capture fine-grained, context-dependent insight quality.

\paragraph{Baseline implementation fidelity.}
The Gemini-row single-shot and LIDA-style baselines are in-house DSPy reproductions of those systems' architectural patterns rather than direct invocations of the published packages; this holds the task model, judge, and sandbox constant at the cost of head-to-head fidelity. The GPT-4.1 row instead invokes the installed lida v0.0.14 package directly and finds an even larger chain advantage, so the Gemini-row LIDA is the more conservative of the two comparisons. The evaluation also lacks a representative general-purpose data-analysis agent adapted with a visualization stage. Consequently, the results establish an advantage over the included executable visualization and prompting baselines, but not over the broader class of agentic exploratory-data-analysis systems.

\paragraph{Prompt-level overfitting.}
The five training domains, $20$-task validation pool, and three independent compilation runs limit the diversity of optimization conditions. Positive gains across $135$ tasks from five unseen domains and across all three runs reduce concerns about test leakage and run-to-run instability, but do not fully rule out prompt-level overfitting. 


\appendix

\section{Supplementary Material}
\label{sec:appendix}

\subsection{IPM Validation Tables}
\label{sec:appendix-ipm-validation}

\begin{figure}[h]
\centering
\includegraphics[width=0.85\linewidth]{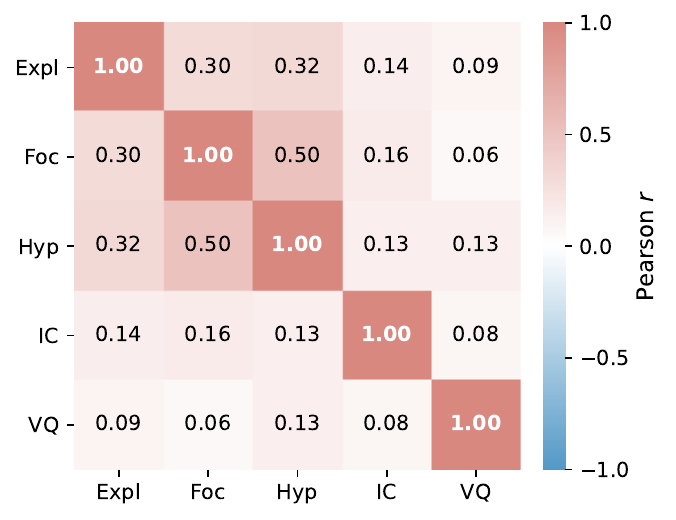}
\caption{Inter-dimension Pearson correlations on 508 complete chains from the original three-condition evaluation: 170 datasets evaluated with single-shot, uncompiled-chain, and LIDA baselines, excluding two failed LIDA evaluations. Mean off-diagonal $\bar{r}=0.19$; max $r=0.50$ (Foc--Hyp). Cronbach's $\alpha$ on the full 5-tuple is $0.39$; the leave-one-out values are $-$Expl $0.37$, $-$Foc $0.29$, $-$Hyp $0.22$, $-$IC $0.37$, $-$VQ $\mathbf{0.44}$ --- \textsc{Visual Quality} is the only dimension whose removal raises $\alpha$, consistent with its capturing variance not covered by the four text-judged dimensions. Expl~=~Exploration Breadth, Foc~=~Focus Relevance, Hyp~=~Hypothesis Rigor, IC~=~Insight Communication, VQ~=~Visual Quality.}
\label{fig:ipm-decoupling}
\end{figure}

This subsection presents the supporting results for §~\ref{sec:ipm-validation}, including per-dimension human–IPM agreement (Table~\ref{tab:human-agreement}), judge self-consistency and cross-family validation (Table~\ref{tab:judge-reliability}), and the inter-dimension correlation heatmap (Figure~\ref{fig:ipm-decoupling}).

\begin{table}[h]
\centering
\small
\setlength{\tabcolsep}{4pt}
\begin{tabular}{lrrrr}
\toprule
\textbf{Dimension} & \textbf{$r$} & \textbf{$\rho$} & \textbf{MAE} & \textbf{h $-$ IPM} \\
\midrule
Exploration Breadth   & $0.78$ & $0.76$ & $2.53$ & $+2.18$ \\
Focus Relevance       & $0.75$ & $0.75$ & $1.54$ & $+1.17$ \\
Hypothesis Rigor      & $0.53$ & $0.51$ & $2.63$ & $+2.24$ \\
Insight Communication & $0.65$ & $0.67$ & $2.53$ & $+2.22$ \\
Visual Quality        & $0.68$ & $\mathbf{0.77}$ & $1.79$ & $\mathbf{-0.04}$ \\
\midrule
\textbf{Composite}    & $\mathbf{0.87}$ & $\mathbf{0.88}$ & $1.88$ & $+1.86$ \\
\bottomrule
\end{tabular}
\caption{Human--IPM agreement on $n=97$ blind-annotated pilot chains. $r$, $\rho$~=~Pearson / Spearman correlations between human and IPM scores on the 1--10 scale (all $p < 10^{-7}$); MAE~=~mean absolute error; \textbf{h $-$ IPM}~=~human-mean minus IPM-mean (the scale-offset bias). Composite row is human overall-quality vs.\ the IPM 5-dim weighted composite. The vision-judged \textsc{Visual Quality} is essentially calibrated to the human ($\Delta = -0.04$); the four text dimensions show a uniform $+1.2$-to-$+2.2$ strictness offset that does not affect rank-based agreement.}
\label{tab:human-agreement}
\end{table}

\begin{table}[h]
\centering
\small
\setlength{\tabcolsep}{4pt}
\begin{tabular}{lrrr}
\toprule
\multicolumn{4}{l}{\textbf{(a) Per-dimension self-consistency ($3$ runs)}} \\
\textbf{Dimension} & \textbf{n} & $\alpha$ & $\bar{r}$ \\
\midrule
Exploration Breadth   & 100 & $0.959$ & $0.959$ \\
Focus Relevance       & 100 & $0.894$ & $0.894$ \\
Hypothesis Rigor      & 100 & $0.918$ & $0.918$ \\
Insight Communication & 100 & $0.934$ & $0.935$ \\
Visual Quality        & \phantom{0}99 & $0.926$ & $0.928$ \\
\midrule
\multicolumn{4}{l}{\textbf{(b) Cross-family validation vs.\ in-family Gemini}} \\
\textbf{Judge} & \textbf{n} & \textbf{Gem / XJ} & $\Delta$ \\
\midrule
GPT-4.1       & 200 & $6.95$ / $6.99$ & $+0.04$ \\
Qwen-VL-Max   & 199 & $6.89$ / $7.57$ & $+0.68$ \\
\bottomrule
\end{tabular}
\caption{IPM judge reliability. \textbf{(a)}~Per-dimension self-consistency on the 100-chain pilot: all five dimensions are independently rescored three times at temperature $0.5$; $n$~=~number of evaluated chains with complete three-run scores, $\alpha$~=~Krippendorff's interval alpha across three judge runs, and $\bar{r}$~=~mean pairwise Pearson $r$. \textbf{(b)}~Cross-family validation against the in-family Gemini~3~Flash judge: ``Gem'' is the in-family mean on the matched subsample, ``XJ'' is the cross-family judge's mean, and $\Delta = \text{XJ} - \text{Gem}$ reports the mean score difference. Both cross-family judges return means at or above Gemini, providing no evidence of in-family score inflation on this matched subsample.}
\label{tab:judge-reliability}
\end{table}

\subsection{IPM Expanded Validation and Close-Pair Evaluation}
\label{sec:appendix-validation-protocols}

\paragraph{Expanded agent-based evaluation.}
Because the 100-chain human study is limited in scale, we expanded the evaluation to 300 generated analysis chains spanning all ten domains. The 300 chains consist of the $100$-chain pilot ($20$ datasets under five regimes) with $200$ additional chains generated from $50$ datasets under four regimes. The resulting condition counts are $70$ \textsc{high}, $90$ \textsc{medium}, $70$ \textsc{low}, and $70$ \textsc{random}. Each annotation record contains the four stage outputs, generated code, execution output, final insight, and rendered Present-stage figure. Annotators assign five integer dimension scores on the 1--10 scale, an independent overall-quality score, and categorical judgments of factual correctness, practical meaningfulness, and effect-size reporting.

An advanced proprietary agent system independently rated every chain three times using the same rubric provided to the human annotators. Its mean ratings show strong agreement with IPM (Pearson $r=0.88$, Spearman $\rho=0.89$, and MAE $=0.93$), while the repeated ratings show high test--retest reliability (overall Krippendorff's $\alpha=0.88$; per-dimension $\alpha=0.63$--$0.91$). A data-analysis expert then audited the agent's judgments of factual correctness, effect-size reporting, and practical meaningfulness, correcting incorrect or ambiguous binary decisions. Based on these expert-corrected labels, practical meaningfulness rises from $0\%$ for \textsc{random}/\textsc{low} outputs to $95.4\%$ for \textsc{high} outputs. These results support the repeatability of IPM and its separation of broad quality regimes, but do not by themselves establish the construct validity of every individual dimension.

\paragraph{Close-pair evaluation.}
To test whether IPM distinguishes outputs of similar quality, agent annotators compared 80 pairs, including regenerated outputs from the same quality regime and outputs with nearly identical IPM scores. The  $80$ pairs are drawn from three sources:

\begin{table}[h]
\centering
\small
\setlength{\tabcolsep}{3pt}
\begin{tabular}{@{}l r p{0.55\linewidth}@{}}
\toprule
\textbf{Source} & \textbf{Pairs} & \textbf{Construction} \\
\midrule
Seed pair & $20$ & Same dataset and prompt; different seed \\
Regeneration & $40$ & Same dataset and quality regime \\
Metric-close & $20$ & Same regime, different datasets, $|\Delta\mathrm{IPM}|<0.5$ \\
\bottomrule
\end{tabular}
\caption{Composition of the close-pair evaluation set.}
\label{tab:close-pair-construction}
\end{table}

Pairs without a rendered Present-stage figure are excluded. Pair order and left--right assignment are randomized, while the quality regime and IPM scores remain hidden during annotation. The blind interface presents the two figures and final insights and requests an A, B, or tie judgment. Preferences are compared with the sign and magnitude of the hidden IPM difference only after annotation. Based on this analysis, differences below approximately $0.8$ IPM points are interpreted as ties. The results indicate that IPM reliably distinguishes outputs when $|\Delta\mathrm{IPM}|>0.8$, whereas smaller differences behave as ties.

\paragraph{Blinding and repeated agent ratings.}
Before annotation, condition labels, automatic IPM scores, tier-bearing chain identifiers, and identifying figure paths are removed or replaced with neutral identifiers. The private identifier mapping is not reopened until a complete annotation pass has been frozen. The first two passes divide the corpus into $20$ blinded packets of $15$ chains, with independently scored packets and different deterministic orderings; the third pass uses $300$ individually masked packets and neutralized figure names. All three passes use the same rubric as the human pilot. For agreement with IPM, the three-run mean for each dimension is combined using the original IPM weights and compared with the primary IPM composite. Test--retest reliability is computed as interval Krippendorff's $\alpha$ across the three passes.

\paragraph{Expert audit.}
The binary criteria are audited against execution-grounded evidence. For factual correctness, the auditor checks every reported number and direction against the chain's executed \texttt{stdout} and, where relevant, the rendered figure. Effect-size reporting requires an explicit magnitude, such as a difference, percentage, ratio, correlation, or standardized effect, rather than a $p$-value alone. Practical meaningfulness is judged independently of factual correctness and distinguishes domain-relevant findings from trivial descriptions. The audit artifacts contain $37$ deliberately enriched disagreement cases: $7$ for meaningfulness, $15$ for effect-size reporting, and $15$ for factual correctness. The auditor records an evidence basis, rationale, and corrected verdict for each case. Because this set concentrates on disputed cases, its annotator-agreement rates are not interpreted as population estimates.

\subsection{Metric Comparison}
\label{sec:appendix-metrics-compare}

Table~\ref{tab:metrics-compare} compares IPM with representative metrics from the three families reviewed in §~\ref{sec:rw-metrics} across five design properties.

\begin{table*}[h]
\centering
\small
\setlength{\tabcolsep}{3pt}
\begin{tabular}{lccccc}
\toprule
\textbf{Metric} & \textbf{\makecell{Multi-\\stage}} & \textbf{\makecell{Vision-\\judged}} & \textbf{\makecell{Faithful-\\ness gate}} & \textbf{\makecell{Insight-\\aware}} & \textbf{\makecell{Spec-\\free}} \\
\midrule
nvBench-style spec match \citep{luo2021nvbench} & \ding{55} & \ding{55} & \ding{55} & \ding{55} & \ding{55} \\
Chart-to-Text BLEU \citep{kantharaj2022c2t} & \ding{55} & \ding{55} & \ding{55} & \ding{55} & \ding{51} \\
CLIP \citep{radford2021learningtransferablevisualmodels} & \ding{55} & \ding{51} & \ding{55} & \ding{55} & \ding{51} \\
LIDA partial VLM eval \citep{dibia-2023-lida} & \ding{55} & partial & \ding{55} & partial & \ding{51} \\
InsightBench rubric \citep{sahu2024insightbench} & text only & \ding{55} & \ding{55} & \ding{51} & \ding{51} \\
\midrule
\textbf{IPM (ours)} & \ding{51} & \ding{51} & \ding{51} & \ding{51} & \ding{51} \\
\bottomrule
\end{tabular}
\caption{Visualization quality metrics by design property. \textbf{Multi-stage}: does the metric score intermediate chain stages rather than only the final figure? \textbf{Vision-judged}: does it consume the rendered image? \textbf{Faithfulness gate}: is there an explicit cap for misleading figures? \textbf{Insight-aware}: does it reward task-specific finding extraction? \textbf{Spec-free}: does it avoid requiring a gold chart specification?}
\label{tab:metrics-compare}
\end{table*}

\subsection{APO Design-Space Summary}
\label{sec:appendix-apo-design}

Table~\ref{tab:apo} consolidates the five APO methods of §~\ref{sec:apo} along their two distinguishing axes (search algorithm and search target).

\begin{table}[h]
\centering
\small
\setlength{\tabcolsep}{4pt}
\begin{tabular}{lll}
\toprule
\textbf{Method} & \textbf{Search algorithm} & \textbf{What is searched} \\
\midrule
MIPROv2        & Bayesian / TPE       & instr.\ + demos \\
COPRO          & Coordinate ascent    & instr.\ only \\
OPRO           & LLM-as-optimizer     & instr.\ only \\
EvoPrompt      & Evolutionary         & instr.\ only \\
Plan-and-Solve & none (template)      & --- \\
\bottomrule
\end{tabular}
\caption{The five APO paradigms compiled and evaluated under identical
splits, metric, and budget. Plan-and-Solve has no learnable parameters
and serves as a structured-template control.}
\label{tab:apo}
\end{table}

\subsection{Stage Instructions and VG-COPRO Algorithm}
\label{sec:appendix-algorithm}

\paragraph{Stage instructions $\theta_t$.} The four seed instructions that VG-COPRO
mutates (taken verbatim from the DSPy signature docstrings; Explore is also shown
in §~\ref{sec:framework}):
\begin{itemize}\itemsep1pt\topsep1pt\leftmargin1.1em
\item \textbf{Explore} ($\theta_1$): \emph{``Given a dataset summary and sample, generate Python visualization code that broadly explores the data landscape with 2--3 overview charts. Identify distributions, relationships, temporal patterns, and outliers. Print key observations and identify non-obvious patterns worth investigating.''}
\item \textbf{Focus} ($\theta_2$): \emph{``Given exploration insights and key patterns, drill down into the most promising pattern. Use groupby, filtering, or segmentation to investigate in depth. Formulate a specific, testable hypothesis.''}
\item \textbf{Test} ($\theta_3$): \emph{``Given a hypothesis from the Focus step, design and execute a statistical test to validate or refute it. Use \texttt{scipy.stats} for appropriate tests. Report $p$-value, effect size, and confidence intervals.''}
\item \textbf{Present} ($\theta_4$): \emph{``Given all prior insights from a full analysis chain, create ONE polished, publication-ready annotated visualization that communicates the key finding. Use annotations, arrows, color, and a headline title that states the finding.''}
\end{itemize}

\paragraph{Algorithm.}
The full pseudocode for VG-COPRO is given in Algorithm~\ref{alg:vg-copro}. Inline comments indicate where each modification (Mod~1--3) alters the COPRO~$+$~OPRO-proposer baseline in §~\ref{sec:apo}.

\begin{algorithm}[h]
\caption{VG-COPRO}
\label{alg:vg-copro}
\small
\begin{algorithmic}[1]
\Require chain $\Pi = (\pi_1, \ldots, \pi_4)$; LODO folds $\{\mathcal{F}_d\}_{d=1}^{5}$; proposer $g$; text/vision judges $J_t, J_v$; iterations $N{=}3$, history $K{=}5$

\For{$\pi \in \Pi$}
  \State $\mathcal{H} \gets \big[(\pi, \mathrm{eval}(\Pi))\big]$
  \For{$t = 1, \ldots, N$}
  
    \State $\triangleright$ \emph{Propose (Mod~1: diagnostic, per-dimension feedback)}
    \State $\pi' \gets g(\mathrm{top}\text{-}K(\mathcal{H}))$;\quad replace $\pi$ with $\pi'$ in $\Pi$
    
    \State $\triangleright$ \emph{Evaluate (Mod~2: LODO folds; Mod~3: vision-aware IPM)}
    \For{$d = 1, \ldots, 5$}
      \State $\bar{s}_d \gets \mathrm{mean}_{x \in \mathcal{F}_d}\, \mathrm{IPM}\!\big(J_t(\Pi(x)) \oplus J_v(\Pi(x).I)\big)$
    \EndFor
    
    \State $\triangleright$ \emph{Aggregate (worst-fold objective)}
    \State $\mathrm{fit} \gets \min_d \bar{s}_d$
    \State append $(\pi', \mathrm{fit})$ to $\mathcal{H}$
    
  \EndFor
  \State $\pi \gets \arg\max_{\pi' \in \mathcal{H}} \mathrm{fit}$
\EndFor

\Ensure compiled chain $\Pi^*$
\end{algorithmic}
\end{algorithm}


\subsection{Optimization Dynamics}
\label{sec:appendix-apo-dynamics}

\begin{figure*}[t]
    \centering
    \includegraphics[width=\textwidth]{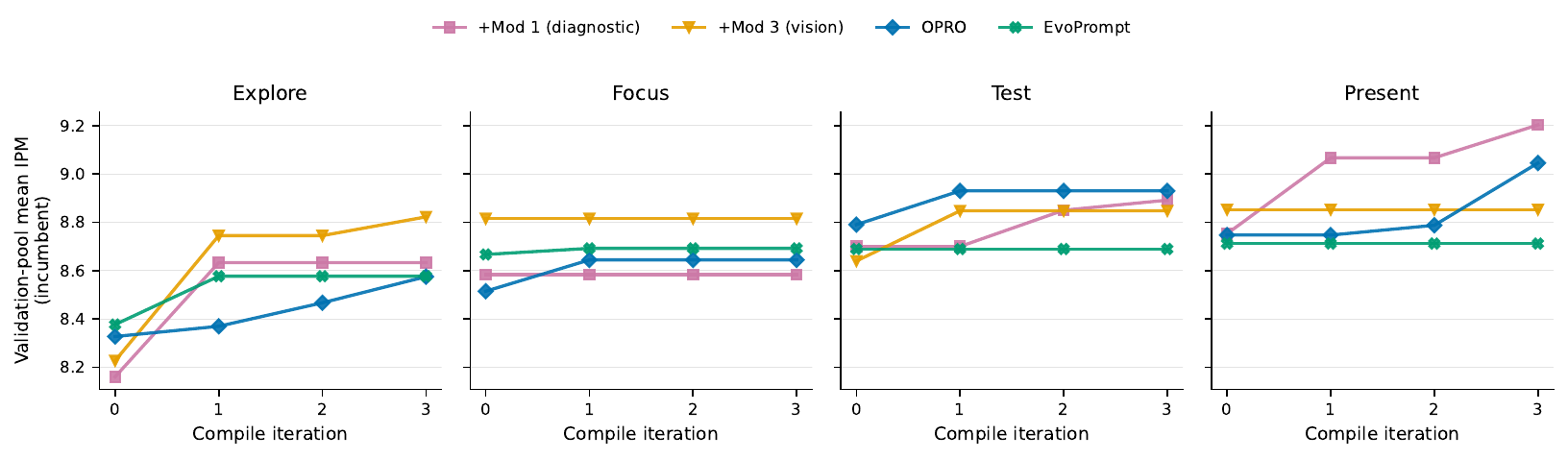}
    \caption{Checkpoint-reconstructed optimization dynamics over three compile iterations. Each panel plots the validation-pool mean of the incumbent selected under each method's own objective. Curves are comparable over iterations within a method, but absolute heights should not be compared across metric variants: OPRO, EvoPrompt, and Mod~1 use four text dimensions, whereas Mod~3 uses the full five-dimensional IPM. VG-COPRO and the isolated LODO variant are omitted because their worst-fold selection objective is not directly comparable to the plotted mean-score selection trajectories; methods without reconstructable validation-mean histories are also omitted.}
    \label{fig:apo-dynamics}
\end{figure*}

Figure~\ref{fig:apo-dynamics} reconstructs the incumbent validation trajectory from the released per-method state checkpoints. Iteration~0 is the seed instruction, and iterations~1--3 are the three candidate-proposal rounds. At each round, the plotted point is the validation-pool mean of the incumbent selected according to that method's own objective. OPRO improves in all four stages, with final stage-wise gains of $0.25$, $0.13$, $0.14$, and $0.30$ validation-score points for Explore, Focus, Test, and Present, respectively. EvoPrompt improves in two stages and is unchanged in the other two. The diagnostic-only and vision-only variants improve in three and two stages, respectively. The frequent plateaus indicate that useful prompt updates are concentrated in particular stages rather than arising uniformly at every iteration. Accordingly, the figure shows that APO can improve stage-level validation scores within individual methods, with gains confined to particular stages and proposal rounds.

These within-objective gains do not necessarily transfer to the multimodal held-out evaluation. The five published APO baselines optimize only textual artifacts from the reasoning chain, whereas the final IPM also evaluates the rendered figure. OPRO provides the clearest example of this mismatch: although its text-only validation trajectory improves in all four stages, its cross-domain full-IPM mean falls from $7.38$ before APO to $7.27$ after optimization (Table~\ref{tab:baselines}). Among chains with complete cross-domain dimension ratings, all five published APO methods also score $0.33$--$0.87$ points lower than pre-APO InsightChain on \textsc{Insight Communication} and $0.56$--$1.19$ points lower on \textsc{Visual Quality}. Thus, improvements under a text-only compile objective do not ensure visually faithful, well-annotated figures that communicate the discovered insight. Consistent with this interpretation, the vision-aware Mod~3 uses the full five-dimensional IPM during compilation and improves held-out cross-domain IPM by $1.23$ points (Table~\ref{tab:apo-ablation}). Together, these results motivate aligning the optimization signal with the multimodal final evaluation, as VG-COPRO does.


\subsection{Complete Qualitative Analysis Chains}
\label{sec:appendix-full-chains}

\begin{figure*}[t]
    \centering
    \includegraphics[width=0.82\textwidth]{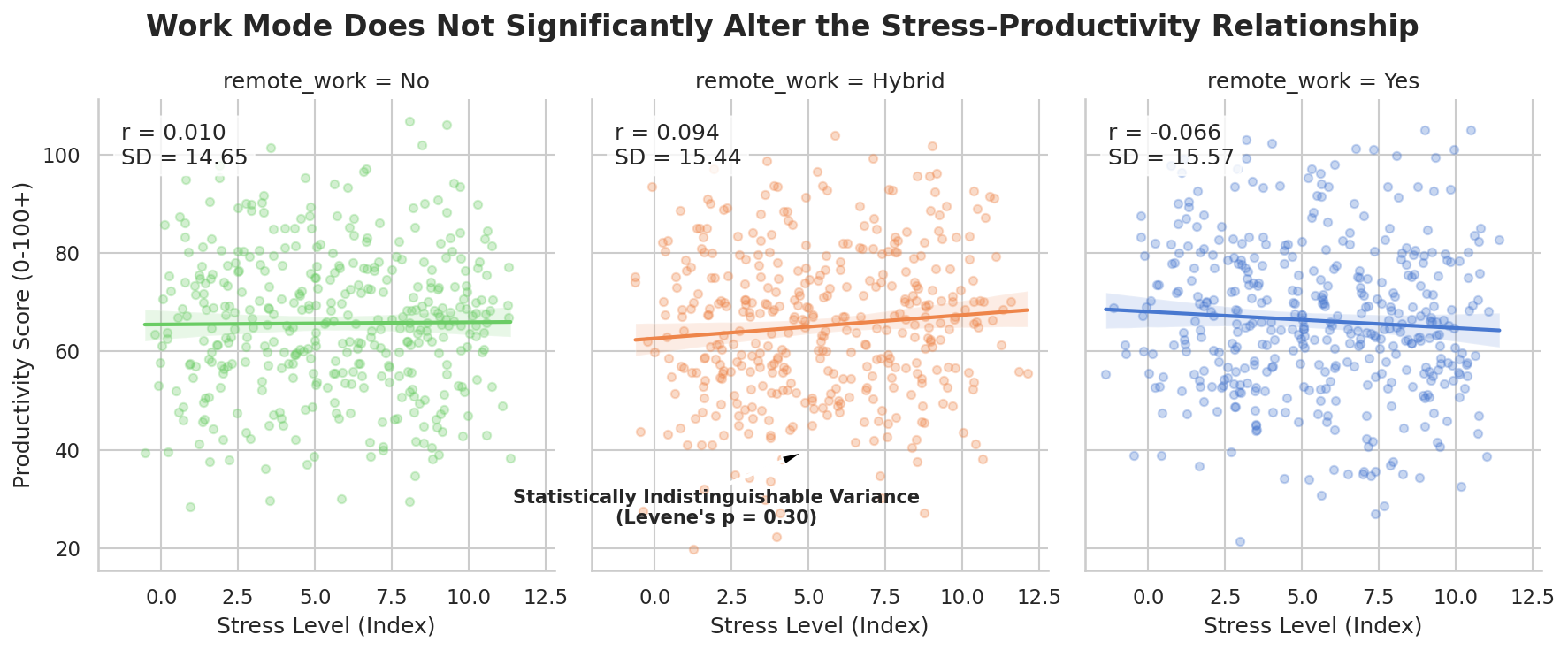}
    \caption{Complete healthcare analysis chain. The example shows the pipeline revising an exploratory impression after formal testing rather than forcing a positive finding.}
    \label{fig:qual-chain-health}
\end{figure*}

\begin{figure*}[t]
    \centering
    \includegraphics[width=0.78\textwidth]{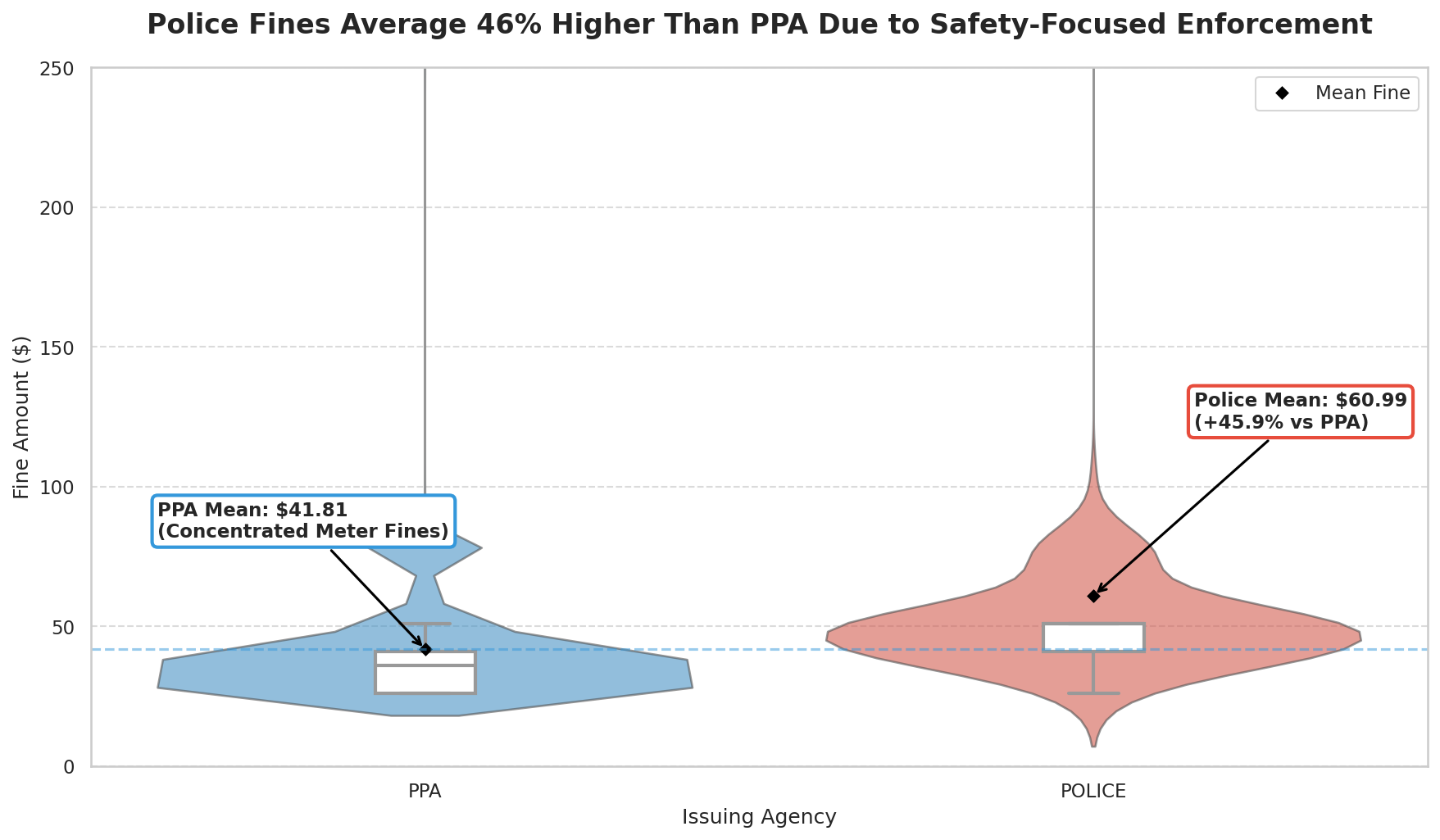}
    \caption{Complete transportation analysis chain. This held-out-domain example progresses from broad enforcement patterns to a quantified agency-level difference with uncertainty and effect size.}
    \label{fig:qual-chain-tickets}
\end{figure*}

To make the analytical depth behind the aggregate scores directly inspectable, Figures~\ref{fig:qual-chain-health}--\ref{fig:qual-chain-reviews} reproduce three VG-COPRO outputs from different domains. The examples cover one in-domain and two held-out cross-domain tasks and include both supported and refuted hypotheses. For readability, we omit the generated Python code but retain the complete semantic progression through Explore, Focus, Test, and Present, including the decision-relevant statistics and caveats. IPM tuples follow the order (\textsc{Exploration Breadth}, \textsc{Focus Relevance}, \textsc{Hypothesis Rigor}, \textsc{Insight Communication}, \textsc{Visual Quality}).

\subsubsection{Healthcare: Mental-Health Survey}

\textbf{Dataset and score.} \texttt{mental\_health\_survey} (healthcare; in-domain test; 1,200 rows). IPM tuple $=(8,9,9,9,10)$; composite IPM $=8.90$. Figure~\ref{fig:qual-chain-health} shows the final visualization.

\paragraph{Explore.} The initial exploration compared stress, productivity, treatment seeking, sleep, and work arrangement. It suggested a weak stress--productivity relationship, higher stress among respondents who sought treatment, and a possible buffering role for sleep, while showing wide productivity variation within every work mode.

\paragraph{Focus.} The chain narrowed the question to whether hybrid work changes the stress--productivity relationship. It hypothesized that hybrid workers would have significantly greater productivity variance than in-office workers and that their stress--productivity correlation would be at least $0.10$ weaker.

\paragraph{Test.} The hypothesis was refuted. Hybrid and in-office productivity standard deviations were $15.44$ and $14.65$, respectively, but Levene's test found no significant variance difference ($p=0.3048$). Their Pearson correlations were $r=0.094$ and $r=0.010$, so the hybrid relationship was slightly stronger rather than weaker; Fisher's $z$ test also found no significant difference ($p=0.2581$). The chain further noted that all correlations were below $|0.10|$, limiting substantive interpretation.

\paragraph{Present.} The final three-panel figure exposes the near-flat regression slopes and comparable dispersions for in-office, hybrid, and remote respondents. Its conclusion is correspondingly negative: work arrangement does not significantly alter either productivity variance or the observed stress--productivity relationship in this sample.

\subsubsection{Transportation: Parking Tickets}

\textbf{Dataset and score.} \texttt{tickets} (transportation; cross-domain test; 50,000 rows). IPM tuple $=(8,9,10,9,10)$; composite IPM $=9.10$. Figure~\ref{fig:qual-chain-tickets} shows the final visualization.

\paragraph{Explore.} The chain found that meter-related violations dominate citation volume, ticketing peaks around late morning and early afternoon, and the Philadelphia Parking Authority (PPA) issues nearly $90\%$ of tickets. Police issue fewer citations but span a wider range of fines.

\paragraph{Focus.} It connected the agency-level disparity to violation composition: PPA activity concentrates on meter expiration and overtime violations, whereas Police activity includes more vehicle-legality and traffic-safety violations. The resulting hypothesis was that the Police mean fine is at least $25\%$ higher than the PPA mean.

\paragraph{Test.} A Welch two-sample $t$-test supported the hypothesis: Police and PPA means were US\$60.99 and US\$41.81, a $45.87\%$ increase and a US\$19.18 difference ($t=25.27$, $p<2.2\times10^{-16}$, Cohen's $d=0.53$; $95\%$ CI for the difference [US\$17.69, US\$20.67]). The chain cautioned that Police fines include high-value outliers and that the PPA sample is almost ten times larger, while noting that the Police median is also higher (US\$51 vs. US\$36).

\paragraph{Present.} The hybrid violin--box plot shows the two full distributions, marks both means, and annotates the $45.9\%$ gap. The final insight attributes the difference to distinct enforcement portfolios while presenting the composition explanation as an interpretation rather than a causal estimate.

\begin{figure*}[t]
    \centering
    \includegraphics[width=0.78\textwidth]{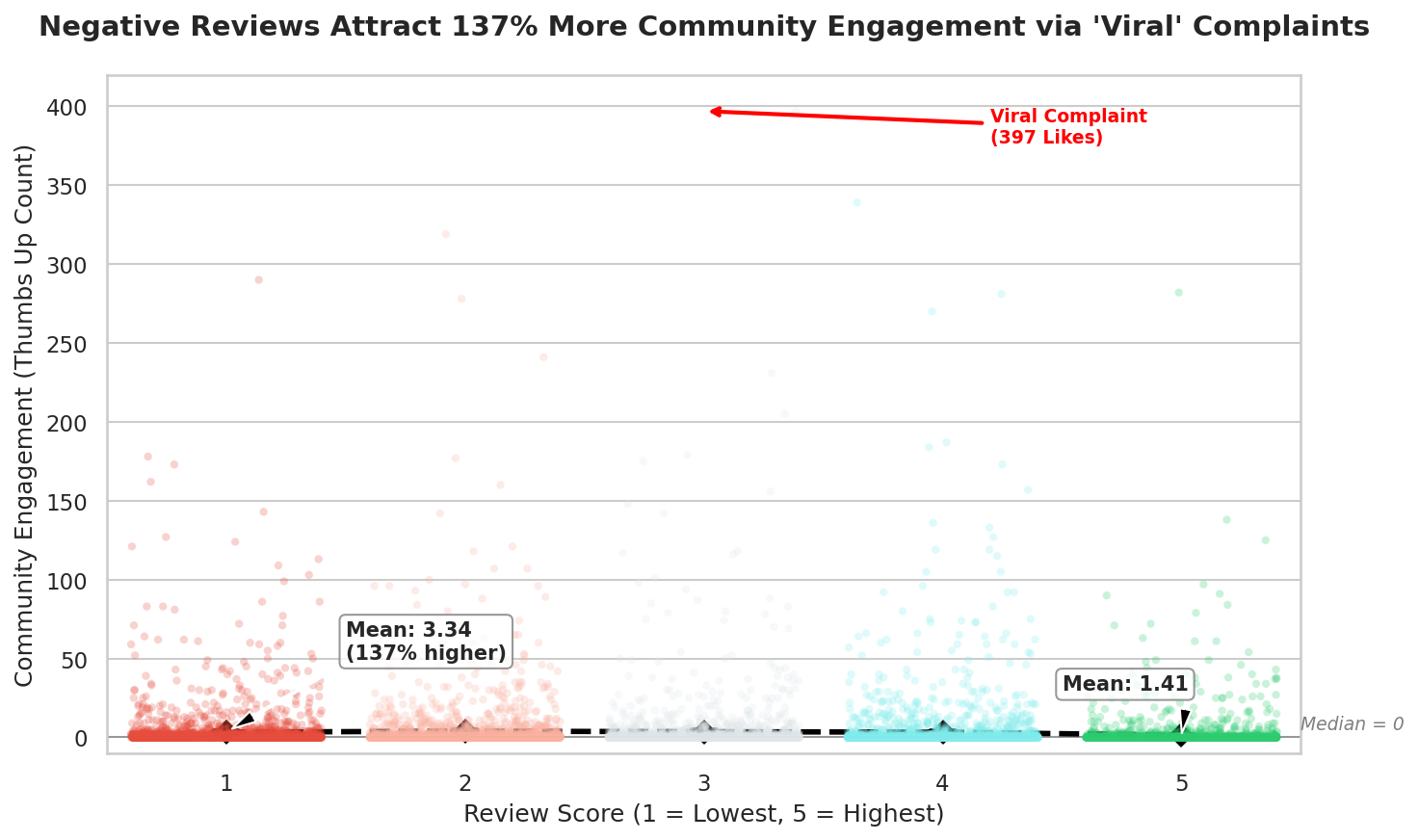}
    \caption{Complete ecommerce analysis chain. This held-out-domain example preserves a nuanced negative decision: the directional effect is significant, but its magnitude does not meet the $200\%$ threshold specified before the Test stage and is dominated by outliers.}
    \label{fig:qual-chain-reviews}
\end{figure*}

\subsubsection{Ecommerce: Google Play Reviews}

\textbf{Dataset and score.} \texttt{google\_play\_store\_reviews\_did43565} (ecommerce; cross-domain test; 14,076 rows). IPM tuple $=(9,9,9,10,10)$; composite IPM $=9.35$. Figure~\ref{fig:qual-chain-reviews} shows the final visualization.

\paragraph{Explore.} The chain identified polarized ratings, selective developer replies to low-score reviews, and substantial variation across applications. It proposed examining whether detailed complaints receive more community endorsement through the \texttt{thumbsUpCount} field.

\paragraph{Focus.} The drill-down found higher mean engagement for one-star reviews but a median near zero at every score, suggesting that a small number of viral complaints drives the difference. The hypothesis deliberately imposed a strong threshold: one-star reviews would receive at least $200\%$ more mean engagement than five-star reviews.

\paragraph{Test.} The thresholded hypothesis was refuted even though the direction was significant. Mean thumbs-up counts were $3.34$ for one-star and $1.41$ for five-star reviews, a $136.9\%$ increase (Welch's $t=4.885$, $p=1.07\times10^{-6}$, Cohen's $d=0.134$). The small effect size and a maximum of 397 thumbs-up votes support the chain's caveat that rare, highly visible complaints drive the mean difference.

\paragraph{Present.} A transparent jitter plot retains the dense mass at zero rather than hiding it behind group averages, annotates the two means, and points to the 397-vote outlier. The final insight therefore distinguishes statistically reliable mean separation from the typical review's near-zero engagement.

\subsection{Annotator Handbook}
\label{sec:appendix-annotator}
\begin{figure*}[p]
    \centering
    \includegraphics[page=1,width=\textwidth]{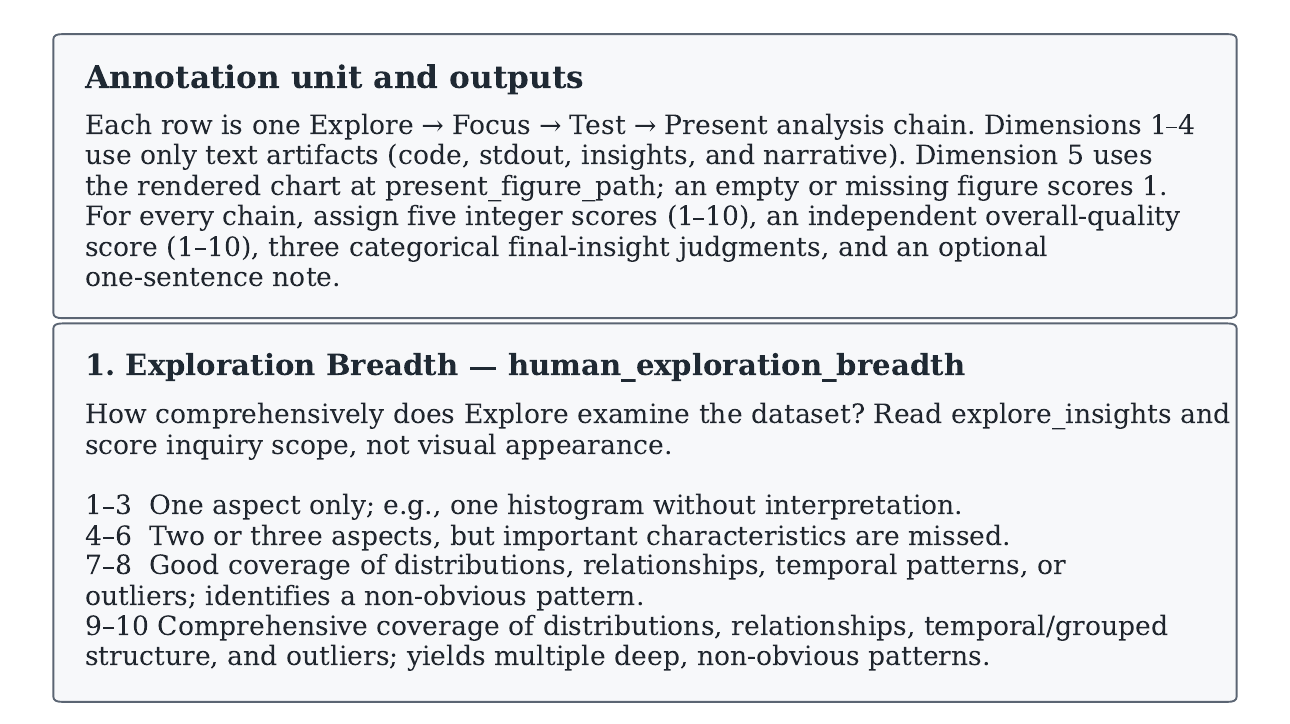}
    \caption{Annotator handbook (1/8): annotation unit, required outputs, and scoring anchors for \textsc{Exploration Breadth}.}
    \label{fig:annotator-handbook-1}
\end{figure*}

\begin{figure*}[p]
    \centering
    \includegraphics[page=2,width=\textwidth]{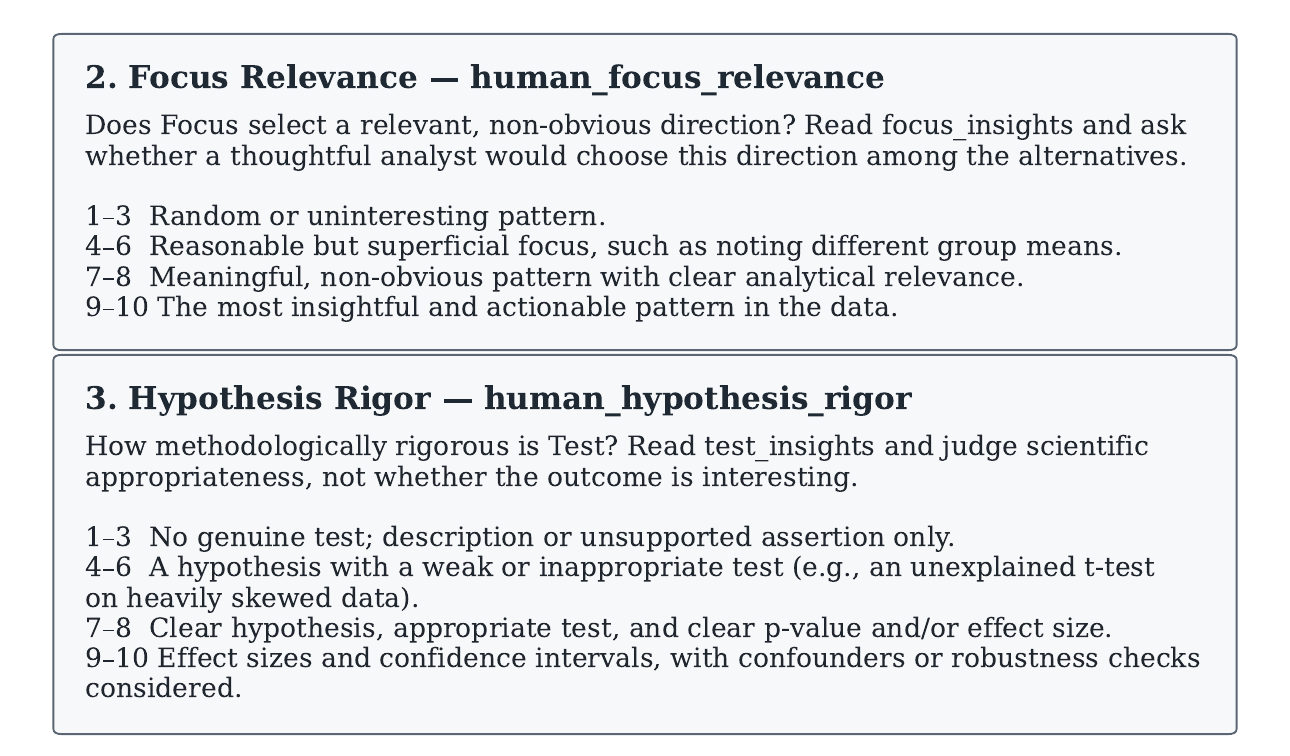}
    \caption{Annotator handbook (2/8): scoring anchors for \textsc{Focus Relevance} and \textsc{Hypothesis Rigor}.}
    \label{fig:annotator-handbook-2}
\end{figure*}

\begin{figure*}[p]
    \centering
    \includegraphics[page=3,width=\textwidth]{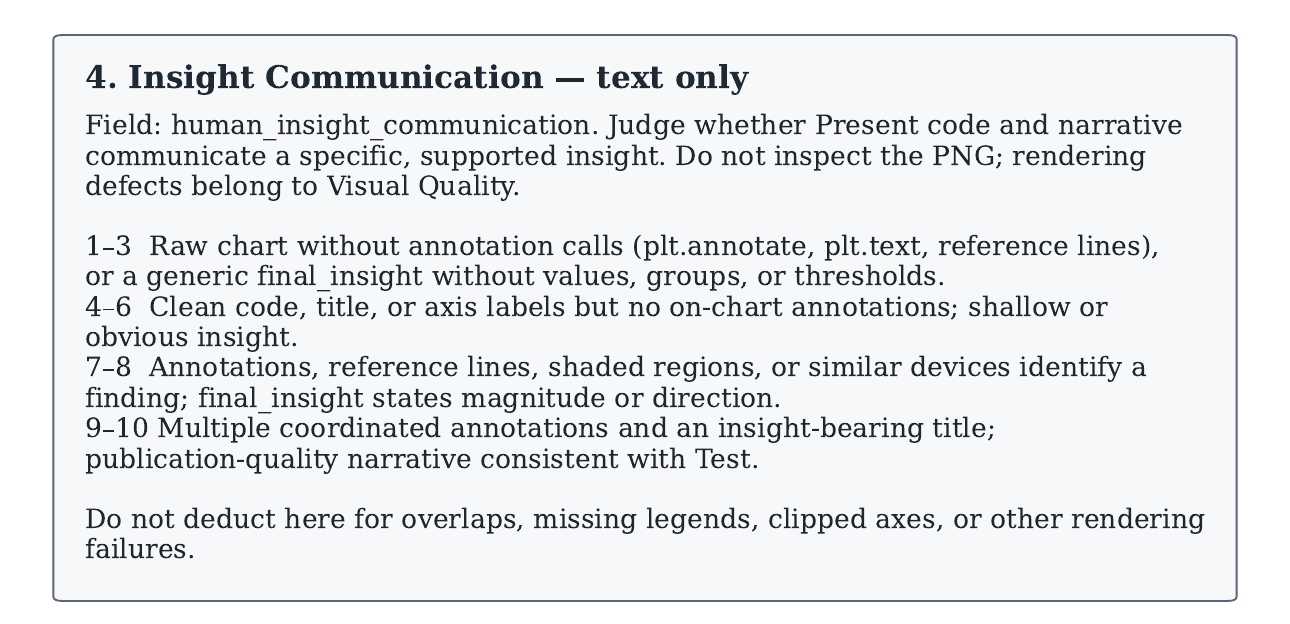}
    \caption{Annotator handbook (3/8): scoring anchors for text-only \textsc{Insight Communication}.}
    \label{fig:annotator-handbook-3}
\end{figure*}

\begin{figure*}[p]
    \centering
    \includegraphics[page=4,width=\textwidth]{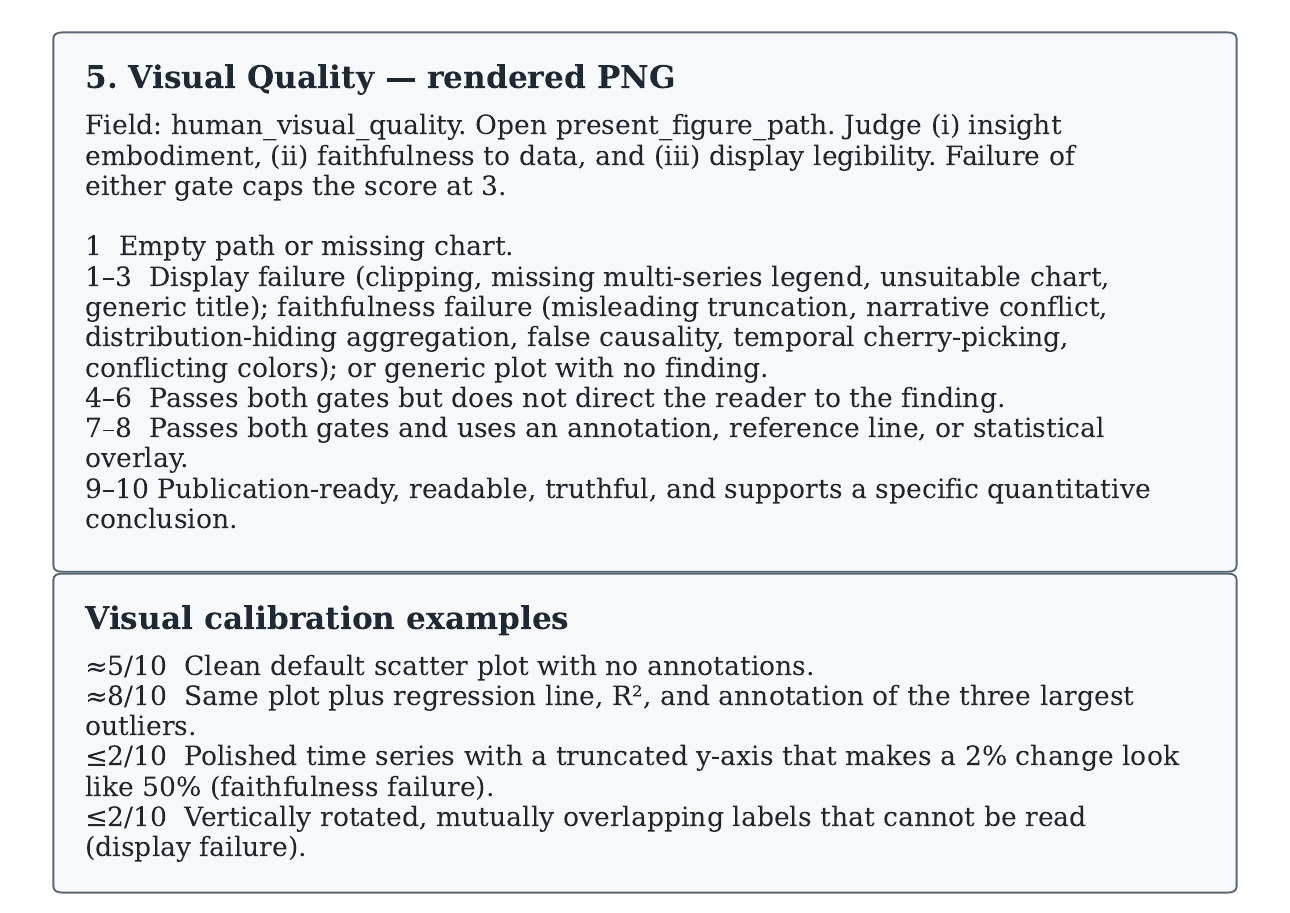}
    \caption{Annotator handbook (4/8): scoring anchors and calibration examples for rendered \textsc{Visual Quality}, including the faithfulness and display gates.}
    \label{fig:annotator-handbook-4}
\end{figure*}

\begin{figure*}[p]
    \centering
    \includegraphics[page=5,width=\textwidth]{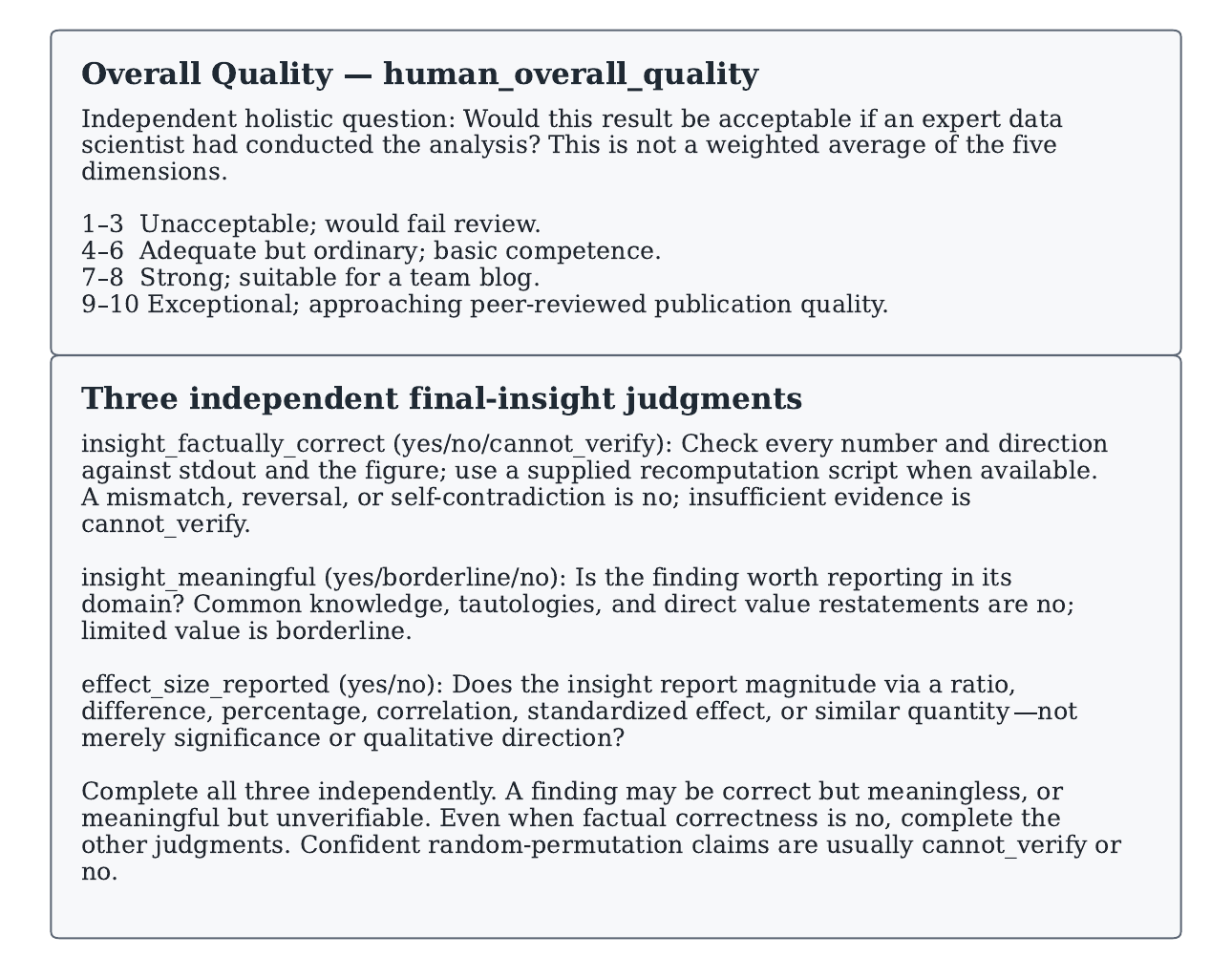}
    \caption{Annotator handbook (5/8): holistic quality and the three independent final-insight judgments.}
    \label{fig:annotator-handbook-5}
\end{figure*}

\begin{figure*}[p]
    \centering
    \includegraphics[page=6,width=\textwidth]{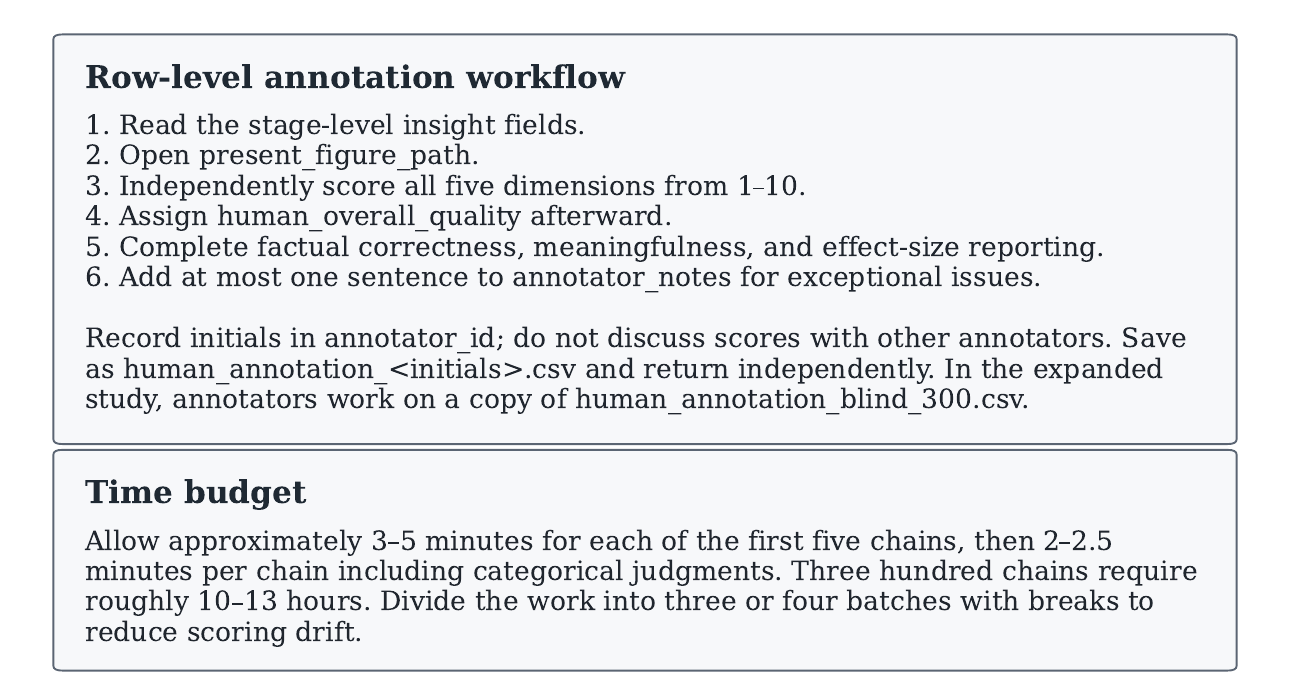}
    \caption{Annotator handbook (6/8): row-level annotation workflow and time budget.}
    \label{fig:annotator-handbook-6}
\end{figure*}

\begin{figure*}[p]
    \centering
    \includegraphics[page=7,width=\textwidth]{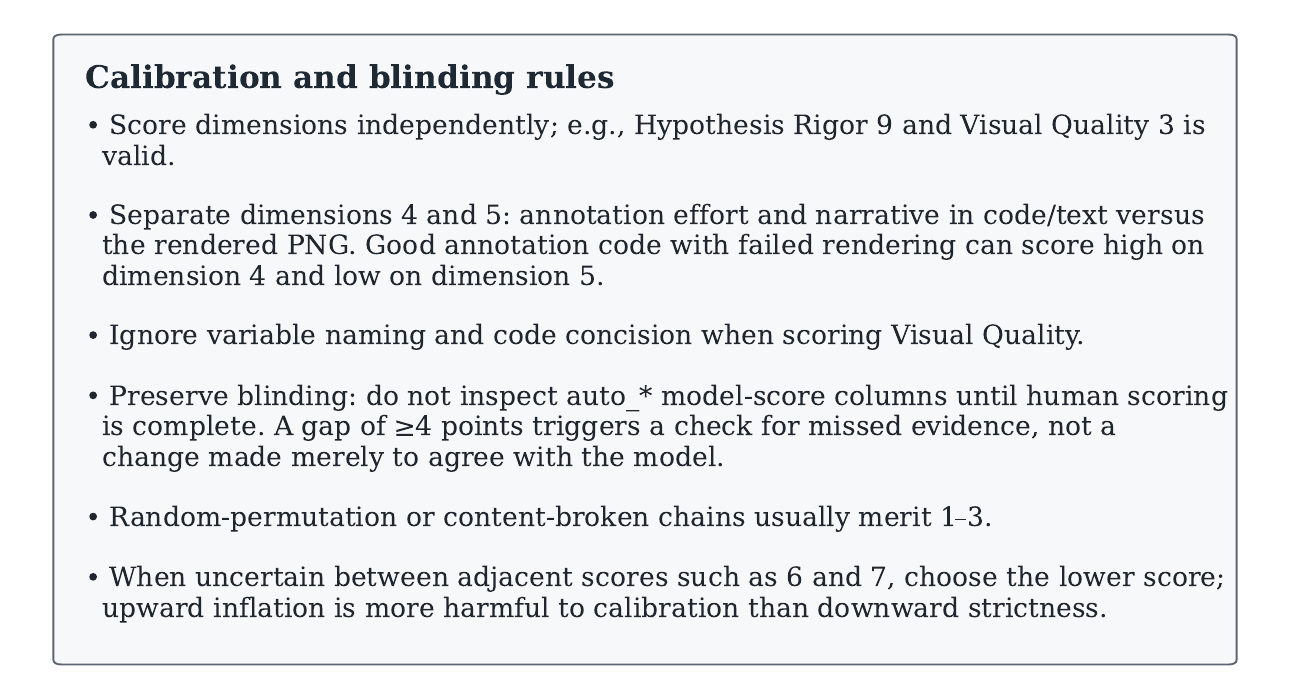}
    \caption{Annotator handbook (7/8): calibration and blinding rules.}
    \label{fig:annotator-handbook-7}
\end{figure*}

\begin{figure*}[p]
    \centering
    \includegraphics[page=8,width=\textwidth]{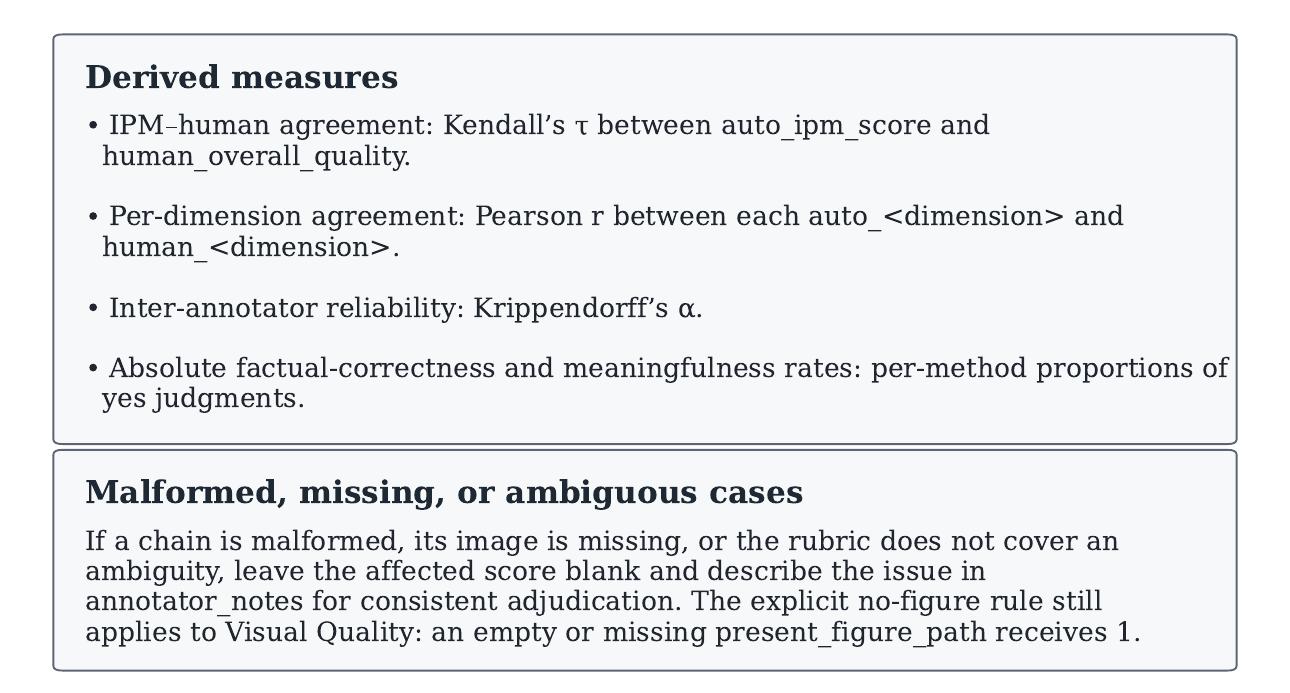}
    \caption{Annotator handbook (8/8): derived agreement measures and handling of malformed, missing, or ambiguous cases.}
    \label{fig:annotator-handbook-8}
\end{figure*}

The complete handbook is reproduced in Figures~\ref{fig:annotator-handbook-1}--\ref{fig:annotator-handbook-8}. Each row is a four-stage analysis chain; dimensions~1--4 are scored from text artifacts, while \textsc{Visual Quality} is scored from the rendered Present-stage figure. Annotators assign five independent 1--10 dimension scores, a separate overall-quality score, and categorical judgments of factual correctness, meaningfulness, and effect-size reporting. Automatic scores remain hidden until the human judgments are complete.

\clearpage

\subsection{Quality-Regime System Prompts}
\label{sec:appendix-regimes}
The five regimes in §~\ref{sec:ipm-validation} (\textsc{High}, \textsc{Medium-a},
\textsc{Medium-b}, \textsc{Low}, \textsc{Random-permutation}) are realized by
four distinct system-prompt sets — \textsc{Medium-a} and \textsc{Medium-b} share
the \textsc{Medium} prompt set but are run under two different random seeds, as
explained in §~\ref{sec:ipm-validation}. The verbatim system prompts below are
taken from \texttt{data\_pipeline/chain\_generator.py}; the user prompt template
(dataset summary and previous-stage context) is shared across regimes.

\paragraph{\textsc{High} (senior data scientist).}
\begin{itemize}\itemsep1pt\topsep1pt\leftmargin1.1em
\item \emph{Explore}: ``You are a senior data scientist with 15 years of
  experience. Perform a thorough exploratory data analysis. Create $3$
  diverse, complementary visualizations (distribution, relationship,
  temporal/categorical); use subplots; examine correlations, outliers,
  distributions, missing patterns; print descriptive statistics; identify $3$
  non-obvious patterns worth deeper investigation.''
\item \emph{Focus}: ``Drill into the MOST non-obvious and impactful pattern.
  Use advanced techniques: groupby$+$agg, pivot tables, conditional filtering.
  Create $2$ visualizations that dissect the pattern from different angles.
  Formulate a specific, quantitative, testable hypothesis.''
\item \emph{Test}: ``Perform rigorous hypothesis testing. Choose the
  appropriate statistical test (t-test, chi-square, correlation, regression);
  report $p$-value, effect size (Cohen's $d$, Cram\'er's $V$, etc.), and
  confidence interval; visualize the test result; mention confounders; state a
  nuanced conclusion (not just `significant').''
\item \emph{Present}: ``Create one polished publication-ready chart. Title
  states the finding as a headline (e.g.\ `Higher X linked to 2.3$\times$
  more Y'). Use \texttt{plt.annotate()} for arrows pointing to key data
  points; add explanatory annotations; professional palette; subtitle with
  the key statistic.''
\end{itemize}

\paragraph{\textsc{Medium-a} / \textsc{Medium-b} (data analyst, two seeds).}
\begin{itemize}\itemsep1pt\topsep1pt\leftmargin1.1em
\item \emph{Explore}: ``Create $2$ standard charts (e.g.\ histogram and
  scatter plot); add titles and axis labels; print basic observations; note
  $1$--$2$ patterns.''
\item \emph{Focus}: ``Pick a reasonable pattern from the exploration; use
  groupby or filtering to investigate; create $1$ chart; state a hypothesis
  based on what you see.''
\item \emph{Test}: ``Use a basic statistical test (e.g.\ $t$-test or
  correlation); report the $p$-value; create a simple visualization; state
  whether the hypothesis is supported.''
\item \emph{Present}: ``Create $1$ chart summarizing your main finding; add
  a descriptive title; use reasonable colors.''
\end{itemize}

\paragraph{\textsc{Low} (beginner, one-line tasks).}
\begin{itemize}\itemsep1pt\topsep1pt\leftmargin1.1em
\item \emph{Explore}: ``Just pick ONE column and make a histogram. Use
  \texttt{plt.hist()}. Print the mean. That's it.''
\item \emph{Focus}: ``Pick any two numeric columns and make a scatter plot.
  Use \texttt{plt.scatter()}. Print correlation. That's it.''
\item \emph{Test}: ``Just compare the mean of one column across groups of
  another. Use \texttt{df.groupby().mean()}. Make a bar chart. That's it.''
\item \emph{Present}: ``Just make a bar chart or pie chart of a categorical
  column. Add a title. That's it.''
\end{itemize}

\paragraph{\textsc{Random-permutation} (content-broken control).}
\begin{itemize}\itemsep1pt\topsep1pt\leftmargin1.1em
\item \emph{Explore}: ``Write a MINIMAL Python script. ONLY do
  \texttt{print(df.describe())}; then make ONE chart:
  \texttt{plt.plot(df.iloc[:,0])}. Do NOT add titles or labels. Do NOT
  analyze anything.''
\item \emph{Focus}: ``Just pick a random column and print value counts. Make
  a bar chart with \texttt{plt.bar()}. No analysis, no labels.''
\item \emph{Test}: ``Just print the correlation matrix:
  \texttt{print(df.corr())}. Make a heatmap with
  \texttt{sns.heatmap(df.corr())}. Nothing else.''
\item \emph{Present}: ``Just do
  \texttt{plt.boxplot()}. No title, no labels, no insight.''
\end{itemize}

\subsection{Dataset Tables}
\label{sec:appendix-datasets}

\begin{figure}[h]
\centering
\includegraphics[width=0.85\linewidth]{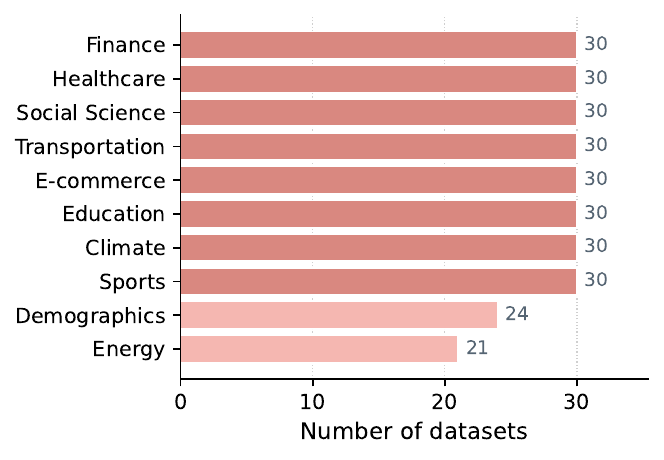}
\caption{Distribution of datasets across the 10 target domains.}
\label{fig:datasets}
\end{figure}

This subsection collects the descriptive summaries of the train/val/test referenced from §\ref{sec:setup}: the property-by-property comparison against prior benchmarks (Table~\ref{tab:benchmark-compare}), source composition (Table~\ref{tab:sources}), per-domain count (Figure~\ref{fig:datasets}), and the leave-five-domains-out split sizes (Table~\ref{tab:splits}).

\begin{table*}[h]
\centering
\small
\setlength{\tabcolsep}{4pt}
\begin{tabular}{l c c c c c c}
\toprule
\textbf{Benchmark} &
\textbf{\makecell{\# Datasets}} &
\textbf{Domains} &
\textbf{\makecell{Multi-\\step}} &
\textbf{\makecell{Cross-\\domain}} &
\textbf{\makecell{Insight\\Metric}} &
\textbf{\makecell{Vision\\Judge}} \\
\midrule
nvBench \citep{luo2021nvbench}           & 750    & Narrow & \ding{55} & \ding{55} & \ding{55}$^{a}$ & \ding{55} \\
Prompt4Vis \citep{li2025prompt4vis}      & subset & Narrow & \ding{55} & \ding{55} & \ding{55}       & \ding{55} \\
Chart-to-Text \citep{kantharaj2022c2t}   & 44k    & --     & \ding{55} & \ding{55} & \ding{55}$^{b}$ & \ding{55} \\
LIDA exemplars \citep{dibia-2023-lida}     & ad-hoc & Few    & \ding{55} & \ding{55} & \ding{55}       & Partial \\
InsightBench \citep{sahu2024insightbench}& 100    & Mixed  & Text-only & \ding{55} & Text            & \ding{55} \\
InsightEval \citep{wang2025insighteval}  & --     & Mixed  & Text-only & \ding{55} & Text            & \ding{55} \\
\midrule
\textbf{InsightChain (ours)} & \textbf{285} & \textbf{10$^{\ast}$} & \textbf{4-step} & \ding{51} & \textbf{5-dim} & \ding{51} \\
\bottomrule
\end{tabular}
\caption{InsightChain vs.\ prior LLM-visualization and insight-quality benchmarks.
$^{a}$Relies on chart-spec matching. $^{b}$Relies on text-based BLEU scores. $^{\ast}$Includes 5 held-out domains.
To our knowledge, InsightChain is the first benchmark to combine a held-out cross-domain partition with a multi-step chain evaluation and a vision-aware judge.}
\label{tab:benchmark-compare}
\end{table*}

\begin{table}[h]
\centering
\small
\setlength{\tabcolsep}{4pt}
\begin{tabular}{lrr}
\toprule
\textbf{Source} & \textbf{\#} & \textbf{\%} \\
\midrule
TidyTuesday (community-curated) & 55  & 19.3 \\
OpenML (keyword search per domain) & 53  & 18.6 \\
Other public CSV (Kaggle / project pages)  & 130 & 45.6 \\
World Bank, FRED \& other macro feeds & 11  & 3.9  \\
UCI ML Repository                  & 8   & 2.8  \\
Yfinance / equity feeds            & 4   & 1.4  \\
sklearn classics                   & 1   & 0.4  \\
\midrule
Synthetic templates (sparse-domain fallback) & 23  & 8.1 \\
\midrule
\textbf{Total} & \textbf{285} & 100.0 \\
\bottomrule
\end{tabular}
\caption{Source composition of the InsightChain experiment datasets. Synthetic templates ($n=23$) are confined to demographics and energy where real public coverage was thinnest after quality filtering.}
\label{tab:sources}
\end{table}

\begin{table}[h]
\centering
\small
\setlength{\tabcolsep}{4pt}
\begin{tabular}{lrl}
\toprule
\textbf{Split} & \textbf{n} & \textbf{Source} \\
\midrule
Train               & 95  & 5 train dom.\ $\times$ 19 \\
Val                 & 20  & 5 train dom.\ $\times$ 4 \\
In-domain test      & 35  & 5 train dom.\ $\times$ 7 \\
\textbf{Cross-domain test} & \textbf{135} & \textbf{5 hold-out dom.\ (all)} \\
\bottomrule
\end{tabular}
\caption{Split sizes. $n$ refers to the number of sampled datasets. The Train split is used for compilation only, and Val for APO candidate selection. The cross-domain test set is the central testbed for the transfer claim of §~\ref{sec:apo-results}.}
\label{tab:splits}
\end{table}

\subsection{Per-Domain Breakdown}
\label{sec:appendix-perdomain}

\begin{figure}[ht]
\centering
\includegraphics[width=\linewidth]{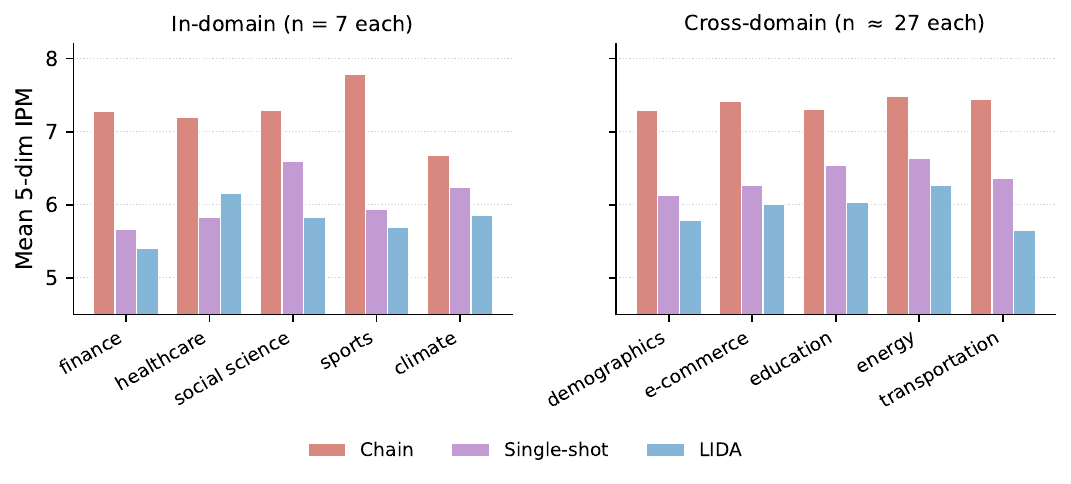}
\caption{Per-domain mean 5-dim IPM under the Gemini~3~Flash task model. InsightChain wins every in-domain training domain and every cross-domain hold-out.}
\label{fig:perdomain}
\end{figure}

Figure~\ref{fig:perdomain} disaggregates the Gemini~3~Flash row of Table~\ref{tab:baselines} by domain. The uncompiled InsightChain outperforms both step-matched baselines across all in-domain training and held-out cross-domain settings. The smallest in-domain gain over single-shot is observed on \textsc{climate} ($+0.44$), where long time-series datasets tend to induce more conservative default plotting behavior, while the largest is on \textsc{sports} ($+1.85$). On the held-out cross-domain evaluation, improvements are more uniform, ranging from $+0.77$ to $+1.16$.

\section{Extended Related Work}
\label{sec:appendix-related}

This appendix complements \S\ref{sec:related-work} by connecting InsightChain, IPM, and VG-COPRO to four adjacent lines of work.

\subsection{Deep Research and Data-Analysis Agents}
\label{sec:appendix-rw-deepresearch}

Deep-research agents interleave reasoning, tool use, and verification rather than answering in one shot \citep{yao2023react, DBLP:journals/corr/abs-2509-06733}. Their treatment of intermediate evidence is especially relevant to InsightChain: Doc-Researcher structures multimodal evidence before reasoning \citep{DBLP:conf/www/DongHYHZLLYWWZL26}, while IE-as-Cache reuses extracted facts across later reasoning steps \citep{DBLP:journals/corr/abs-2604-14930}. InsightChain applies this principle specifically to executable analysis: its context $C_{<t}$ (Eq.~\ref{eq:inputs}) passes computed findings, not merely dialogue history, between stages. Retaining accumulated context can increase inference cost; AsymSpec addresses this complementary systems problem by allowing a lightweight drafter to access the full context while a larger verifier operates on a compressed view \citep{liang2026asymspeccontextasymmetricspeculativedecoding}. Prior work on persistent agent memory \citep{DBLP:journals/corr/abs-2504-15965} and step-level rating and refinement \citep{DBLP:journals/corr/abs-2602-07773} provides complementary foundations for state propagation and stage-specific feedback. Data Interpreter similarly plans and repairs code-executing subtasks \citep{hong2024datainterpreter}, whereas DiscoveryBench evaluates hypothesis search \citep{majumder2024discoverybench} and Multimodal DeepResearcher produces text--chart reports \citep{yang2025multimodaldeepresearcher}. InsightChain instead treats a statistically supported visualization as the final deliverable and evaluates both the analytical path and the rendered figure.

\subsection{Insight Mining and Insight-to-Visualization}
\label{sec:appendix-rw-insight}

Classical insight-mining systems enumerate and rank predefined patterns \citep{ding2019quickinsights, ma2021metainsight}, while storytelling systems compose those patterns into fact sheets or narratives \citep{wang2020datashot, shi2021calliope}. LLM-based systems make exploration more flexible: InsightPilot places an LLM over an insight engine \citep{ma2023insightpilot}, and InsightBench evaluates open-ended insight generation \citep{sahu2024insightbench}. InsightChain combines these directions but grounds each free-form stage in execution: Explore proposes broad candidates, Focus selects hypotheses, Test validates them, and Present renders the result. IPM consequently evaluates whether the final figure communicates the validated finding, rather than merely whether an insight was stated or a chart was produced.

\subsection{LLM-as-Judge Evaluation of Visualizations and Multi-Step Reasoning}
\label{sec:appendix-rw-eval}

IPM builds on reference-free, rubric-based LLM evaluation \citep{liu2023geval}. Systematic analysis of LLM dialogue evaluators shows why agreement with human judgments must be measured rather than assumed \citep{DBLP:conf/aaai/ZhangDCZL24}; adversarial analysis of NLG evaluators further motivates our discriminability controls \citep{DBLP:conf/acl/ChenZLDT024}. VisEval combines automated checks with a vision-language judge for visualization outputs \citep{chen2025viseval}, but assesses query-conditioned chart validity and readability. IPM instead asks whether the final chart faithfully communicates a finding established by the preceding executable stages, while four rubric dimensions separately assess those stages. We therefore report human agreement, repeatability, cross-family robustness, and a content-broken \textsc{random-permutation} control in \S\ref{sec:ipm-validation}.

\subsection{Automatic Prompt Optimization}
\label{sec:appendix-rw-apo}

Automatic prompt optimization searches instructions using scored candidates or textual critiques \citep{zhou2023ape, pryzant2023protegi}. Adjacent adaptation methods operate below the prompt layer: alignment approaches use model-derived rewards or teacher--student fine-tuning \citep{DBLP:conf/aaai/0020C0TGT025, DBLP:conf/emnlp/0020TCSTJ024}, while decoding-time steering combines local and global guidance without changing the task prompt \citep{DBLP:journals/corr/abs-2507-04756, DBLP:journals/corr/abs-2603-16219}. These methods require training or decoding access, whereas prompt optimization applies to black-box APIs. DSPy compilers extend prompt optimization to multi-module programs \citep{khattab2024dspy, opsahl2024miprov2}, but the methods considered in \S\ref{sec:apo} rely on text-only feedback. VG-COPRO retains COPRO's per-predictor search and OPRO-style proposal, while using IPM to supply dimension-specific critique of the full executed pipeline, including the rendered figure. Its leave-one-domain-out worst-case objective \citep{sagawa2020groupdro} further selects prompts for transfer rather than average validation performance. Thus, the novelty is not prompt search itself, but adapting its feedback and selection objective to a multi-stage, visually evaluated analysis program.

\end{document}